\RequirePackage{fix-cm}
\documentclass[twocolumn]{svjour3}          % twocolumn
\smartqed  % flush right qed marks, e.g. at end of proof
\usepackage{graphicx}

\usepackage{algorithm}
\usepackage{algorithmic}
\usepackage{bm}
\usepackage{caption}
\usepackage{comment}
\usepackage{seqsplit}
\usepackage{url}

\renewcommand{\paragraph}[1]{\vspace{1mm} \noindent {\bf #1} \hspace{1.5mm}}

\newcommand{\etal}{et al.}
\usepackage{color}
\newcommand{\yoshiapr}[1]{\textcolor{black}{#1}}% after proofreading
\newcommand{\yoshinew}[1]{\textcolor{black}{#1}}% after AAAI

\newcommand{\yoshi}[1]{\textcolor{black}{#1}}
\newcommand{\kaw}[1]{\textcolor{black}{#1}}
\newcommand{\rk}[1]{\textcolor{black}{#1}} %after second proofreading
\newcommand{\yoshiaapr}[1]{\textcolor{black}{#1}}%after the second proofreading
\newcommand{\ari}[1]{\textcolor{black}{#1}}%ari先生からのコメントを反映
\newcommand{\arm}[1]{\textcolor{black}{#1}}%英文校正2
\newcommand{\yoshifinal}[1]{\textcolor{black}{#1}}%最後にまた変えたくなった箇所
\newcommand{\yoshijj}[1]{\textcolor{black}{#1}}%ジャーナル版2
\newcommand{\yoshij}[1]{\textcolor{black}{#1}}%ジャーナル版
\newcommand{\armj}[1]{\textcolor{black}{#1}}%ジャーナル版 英文校正

\newcommand{\yijcv}[1]{\textcolor[rgb]{0.0,0.0,0.0}{#1}}%IJCV版
\newcommand{\yyijcv}[1]{\textcolor[rgb]{0.0,0.0,0.0}{#1}}%IJCV版

\newcommand{\aijcv}[1]{\textcolor[rgb]{0.0,0.0,0.0}{#1}}%IJCV版 英文校正

\begin{document}

% Differentiating Objects by Motion: 
%\\ Joint Detection and Tracking of 
%Small Flying Objects
    \title{Finding a Needle in a Haystack:
    Tiny Flying Object Detection in 4K Videos using  a Joint Detection-and-Tracking Approach % an idea for new title, 2019/11/24 yoshi
    %\title{Tiny Flying Object Detection in 4K Videos using Motion Cues: A Joint Detection-and-Tracking-based Approach}
}


\author{Ryota~Yoshihashi \and
        Rei~Kawakami \and
        Shaodi~You \and
        Tu~Tuan~Trinh \and
        Makoto~Iida \and
        Takeshi~Naemura
}


\institute{Ryota~Yoshihashi \and Rei~Kawakami 
              \and Tu~Tuan~Trinh \and Makoto~Iida \and Takeshi~Naemura \\at
              The University of Tokyo, 7-3-1, Hongo, Bunkyo-ku, Tokyo \\
              \email{\{yoshi, rei, tu, iida, naemura\}@nae-lab.org}           %  \\
           \and
           Shaodi~You\at
              University of Amsterdam, Postbox 94323, 1090 GH Amsterdam
               \email{s.you@uva.nl}
}

\date{Received: date / Accepted: date}

\maketitle

\begin{abstract}
%\armj{While generic object detection has improved significantly with the advent of rich feature hierarchies from deep nets, detecting small \armj{appearing} objects with poor visual cues has remained challenging.} 
Detecting tiny objects in a high-resolution video is challenging because the visual information is little and unreliable. Specifically, the challenge includes very low resolution of the objects, MPEG artifacts due to compression and a large searching area with many hard negatives.
Tracking is equally difficult because of the unreliable appearance, and the unreliable motion estimation.
Luckily, we found that by combining this two challenging tasks together, there will be mutual benefits.
Following the idea, in this paper, we present a neural network model called the {\it Recurrent Correlational Network}, where detection and tracking are jointly performed over a multi-frame representation learned through a single, trainable, and end-to-end network. 
\yoshij{The framework exploits} \armj{a} convolutional long short-term memory network for learning informative appearance changes for detection, while \armj{the} learned representation is shared in tracking for enhancing its performance. In experiments with datasets containing images of scenes with small flying objects, such as birds and unmanned aerial vehicles, the proposed method yielded consistent improvements in detection performance over deep single-frame detectors and existing motion-based detectors. Furthermore, our network performs as well as state-of-the-art generic object trackers when it was evaluated as a tracker on \armj{a} bird \armj{image} dataset.

\keywords{Detection \and tracking \and tiny-object detection in video \and bird detection \and motion}
\end{abstract}

\section{Introduction}
\begin{figure*}[t]%
%\begin{figure*}[t]%
  \begin{center}
     \hspace{-2mm} \includegraphics[width=500pt]{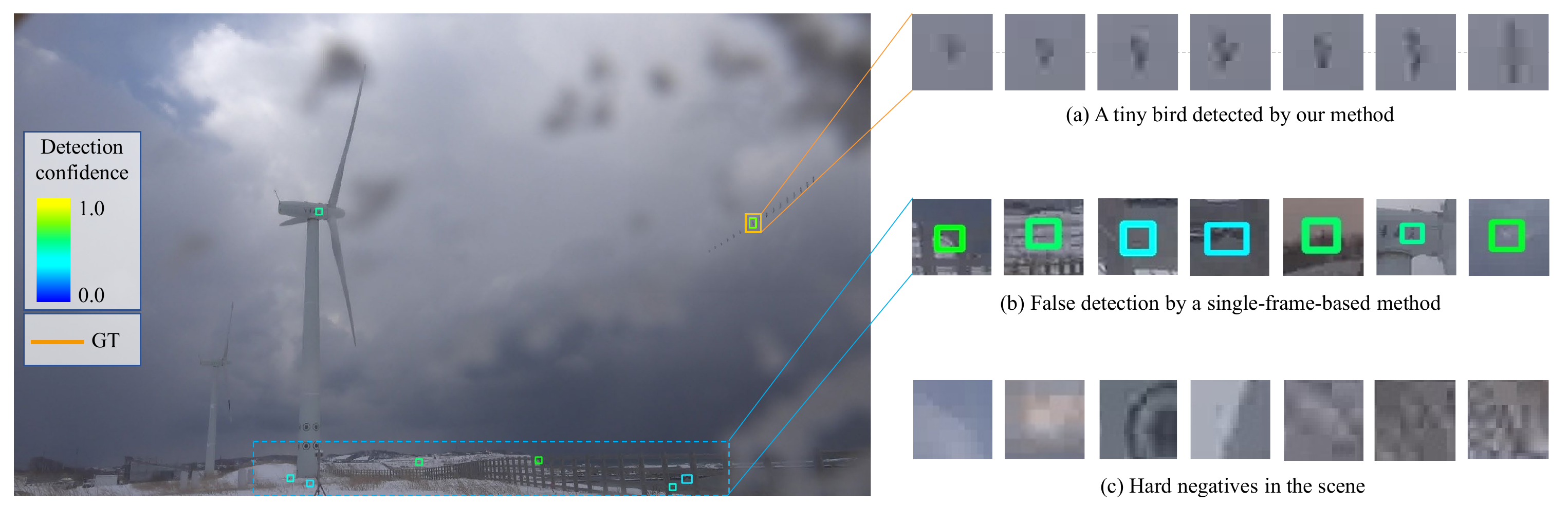}
  \end{center}
  \vspace{-3mm}
  \caption{ \yyijcv{The challenge of tiny flying object detection in high-resolution videos. Multiple frames are overlaid. Apparently small birds in the wild are hard to detect for machines and even for human eyes. In this example, snow accretion on the lens make the detection even harder. In such an environment, motion cues of the objects are beneficial for discriminating target objects from others. Our methods can detect lowly-visible birds by incorporating motion cues, while single-frame-based detectors are prone to miss tiny birds and misdetect backgrounds. Best viewed in zoom in.}}
  %\caption{ Importance of multi-frame information for recognizing apparently small \ari{flying} objects \yoshiaapr{(birds in these examples)}. While \armj{the} visual features in single frames are vague and limited, multi-frame information, including \rk{deformation and pose changes}, provides better clues with which to recognize birds. \yoshifinal{To extract such useful motion patterns, tracking is necessary for compensating \armj{the} translation of objects, but the tracking itself is a challenge due to the limited \armj{amount of} visual information.} }
  \label{fig:motion}
\end{figure*}

\begin{comment}
Detection of visually small objects is often required in wide-area surveillance~\cite{collins2000system,coluccia2017drone,rozantsev2017detecting}.
\yoshijj{Owning to recent \armj{developments} of high-resolution cameras and
automatic object recognition techniques,
wider-area surveillance \armj{of} smaller objects has become possible.
\armj{This capability has a number of applications}, such as \armj{in} bioecological investigations of wild birds or monitoring of unmanned aerial vehicles (UAVs).}
\end{comment}

\yijcv{Recent object detection techniques \aijcv{have} achieved spectacular progress \aijcv{and} dramatically broadened
\aijcv{the range of} applications, \aijcv{to ones} such as traffic monitoring, intelligent security cameras, or biometric authentication.
While many of the applications are designed for indoor or urban environments, 
another interesting direction is to {\it go wild}: examining methods' real-world usability
in natural environments that are harder to control.
For example, ecological investigations of wild birds are still largely conducted by manpower.
Can computer vision be \aijcv{an} aid for bird investigations? --- Unfortunately, current object detectors are
no match for humans in spotting \aijcv{tiny-appearing} birds in wide landscapes.
To enable such applications, object detectors have to be capable of detecting tiny flying objects.}

%\armj{Here, surveillance of small flying objects} requires tailored object detectors, rather than ones for generic \armj{ones};
\yijcv{The problem of {\it tiny flying object detection} \aijcv{presents} several challenges \aijcv{for} object detectors.
First, flying objects often appear \aijcv{in the sky} far from cameras and look small in images, \yyijcv{even when we use high-resolution cameras}. 
Second, the visual variances of flying objects are large due to their various flying directions, poses, and species.
Third, due to \aijcv{high} speed of flying objects, \aijcv{the} image quality of \aijcv{the} object regions tends to be degraded by
motion blur and MPEG compression artifacts.}
\yyijcv{Figure~\ref{fig:motion} shows an example frame from our bird-monitoring setup, including the above-mentioned difficulties. }
\yijcv{Although recent deep object detectors perform impressively well on generic object benchmarks, they are nevertheless insufficient for detecting tiny objects in such hard settings.}
\yijcv{In modern detectors, rich visual representations \aijcv{extracted} by deep convolutional networks (convnets)~\cite{girshick2014rich} are the key workhorses to boost their performance.
The convnets \armj{that have been} pre-trained on a large-scale still-image dataset~\cite{ILSVRC15}
can encode various visual features such as shapes, textures, and colors.
However, for small objects, rich visual features are of limited use by themselves because of the above challenges.}

For detecting \yyijcv{tiny} objects with low visibility, motion, namely the changes in their temporal appearance over a longer time frame, 
may offer richer information than appearance at a glance.
%\yijcv{In this paper, we tackle detection of lowly-visible small object by exploiting motion cues.}
As shown in Fig.~1a, 
%, birds are much easier to identify when multiple frames are available, \armj{even though} their appearance features \armj{may be very} limited. 
with multiple correctly tracked frames, one may tell with more confidence the tracked object is a bird.
However, extracting reliable motion cues is equally difficult as detection because of the limited visual information and the drastic appearance change over time. In our example, the drastic appearance change are mainly from bird-flapping, occlusion from trees, giant \yyijcv{wind turbines}, snow, cloud, etc.
%However, it remains unclear how to learn motion features that are powerful enough to differentiate small \armj{flying} objects.
%\yoshijj{The difficulty comes from the entangled nature of videos;
%\armj{when} moving objects appear, \armj{the} temporal changes of \armj{the} video frames 
%include \armj{the} objects' translations \armj{as well as} appearance changes.
%While the \armj{translations} may \armj{not be so} useful, the \armj{appearance changes} include part-centric motions or deformations that often encode strong class-specific patterns, such as flapping of \armj{wings}, which
%\armj{may} be \armj{very} useful \armj{for detection}.
%When \armj{the translations are large}, such deformation patterns manifest \armj{themselves} as residuals only after the objects' translation is properly compensated.
%Thus, to extract the discriminative  patterns, we first need to disentangle \armj{the} translations and \armj{the} deformations by estimating the translations.}
To be more specific, these challenge makes both dense optical flow estimation and sparse tracking infeasible. For dense optical flow estimation, computing dense flow on a 4K video is  extremely slow. It will also likely smooth out a tiny object. As for sparse tracking, although it is efficient to work on high solution video, the tracker learned on tiny and less informative object will suffer from a similar problem as detection.

\yoshij{In this paper, to effectively exploit motion cues, we tackle these challenges of tiny object detection and tracking \armj{together} in a mutually beneficial manner.}
%In this paper, we present a method that exploits motion cues for small object detection. 
%Although we utilize learnable pipelines based on convolutional and recurrent networks, 
Our key idea is \kaw{letting the network \yoshiaapr{focus on} informative deformations such as flapping of wings to differentiate target objects for detection, while removing less useful translations~\cite{park2013exploring} by simultaneously tracking them with the learned visual representation.}
\yijcv{This compensation of translations is especially important to detect tiny flying objects,
because their displacement is relatively large compared to their size.}
%\arm{To make this possible,}  our framework performs joint detection and tracking. 
%ToDo: rewrite around here 2019/12/11
Our framework utilizes learnable pipelines based on convolutional and recurrent networks
to learn a discriminative multi-frame representation for detection,
while it also enables correlation-based tracking over its output. Tracking is aided by the shared representation afforded by the training of the detector, and the overall framework is simplified \arm{because there are fewer} parameters to be learned.
\yoshiapr{We refer to the pipeline as {\it \armj{a} Recurrent Correlational Network}.}

\yoshij{\armj{Regarding the range of application, we mainly focus on}  single-class, small object detection in videos 
targeting birds~\cite{trinh2016} and unmanned aerial vehicles (UAVs)~\cite{rozantsev2017detecting}.
\armj{The need for detecting} such objects \armj{has grown} with the spread of commercial UAVs, but the \armj{generic} single-image-based detectors are severely challenged by the low resolution and visibility of \armj{these} targets for the purposes.} 
\yoshij{Our experimental results} show \armj{that our network consistently outperforms } single-frame baselines and previous multi-frame methods. 
When evaluated as a tracker, \armj{our network} also outperforms 
existing hand-crafted-feature-based and deep generic-object trackers \armj{on a bird video dataset}.
\yoshij{\armj{In addition,} we  evaluate our method in pedestrian detection~\cite{dollar2012pedestrian}, a task of more general interest in the vision community, and \armj{we found it to be as capable as} the latest pedestrian-specific detectors.}

\yoshij{Our contribution is three-fold. 
\yijcv{First, we indicate that usage of motion cues \aijcv{is} highly effective in tiny object detection, although
the existing literature of tiny object detection took single-image-based approaches~\cite{chenr,hu2016finding,li2017perceptual}.}
Second, we design a novel \armj{network, called the} {\it Recurrent Correlational Network}, that jointly performs object detection and tracking \armj{in order to} \yijcv{effectively extract motion features from tiny moving objects.}
%bridge the gap between conventional single-image-based object detectors and correlation-based class-agnostic object trackers.
%Second, using the RCN framework, we show how joint detection and tracking are mutually beneficial.
%that is, the detector enriches the tracker by giving semantic knowledge to learned representations, and the tracker gives temporal attention to the detector.
Third, our network \armj{performs better} both as a detector and as a tracker \armj{in comparison with} existing methods \armj{on} multiple flying-object datasets, which indicates the importance of motion cues in these domains. 
%RCN outperforms single-frame baselines, score-averaging baselines, and existing multi-frame methods in flying-object datasets, which indicates the importance of motion cues in these domains.
%It is also accurate when evaluated as a separate tracker in the dataset where class-specific detectors can be trained. The proposed network outperforms existing trackers based on various hand-crafted features, and performs slightly better or on par \arm{with} convnet-based trackers. 
Our results \armj{shows} prospects toward domain-specific multi-task representation learning \armj{and} applications that 
generic detectors or trackers do not directly generalize \armj{to}, for example, flying-object surveillance.
\yyijcv{Supplementary videos are available at \url{http://bird.nae-lab.org/tfod4k}.}}
%The relevant code and data will be published upon acceptance of this paper.}

%-----------------------------------------------------------------------------------------------
\section{Related work}
\noindent {\bf Small object detection} \hspace{1.5mm}
\yoshiapr{Detection of small (in appearance) objects has been tackled in \arm{the} surveillance community~\cite{collins2000system}, and it has attracted much attention \arm{since the advent} of UAVs~\cite{coluccia2017drone,schumann2017deep}. There has also been studies focusing on particular objects, for instance, \rk{\ari{small} pedestrians~\cite{bunel2016detection} and faces \cite{hu2016finding}. Recent studies have tried to detect small common objects in a generic-object detection setting~\cite{chenr,li2017perceptual}. These studies have used scale-tuned convnets with moderate depths and a wide field of view, but despite its importance, they have not incorporated motion.} }
\vspace{1mm}

\noindent {\bf Object detection in video} \hspace{1.5mm}
Having achieved significant success with generic object detection on still images~\cite{girshick2014rich,girshick2015fast,ren2015faster,liu2016ssd,dai2016r,redmon2016yolo9000}, \ari{researchers have} begun examining how to perform generic object detection efficiently on videos~\cite{ILSVRC15}. \yoshiapr{The video detection task poses new challenges, such as how to process voluminous video data efficiently and how to handle the appearance of objects that differ from those in still images because of rare poses ~\cite{feichtenhofer2017detect,zhu2017flow}.} The most recent studies have tried to improve the detection performance; examples include T-CNNs~\cite{kang2017t,kang2016tubelets} that use trackers for propagating high-confidence detections, and deep feature flow~\cite{zhu2016deep} and flow-guided feature aggregation~\cite{zhu2017flow} that involve feature-level smoothing using optical flow. \yoshinew {One of the closest idea to ours is joint detection and bounding-box linking by coordinate regression~\cite{feichtenhofer2017detect}. \yoshiapr{However, these models, \arm{which have been entered} in ILSVRC-VID competition, are more like models with temporal consistency than ones that understand motion. Thus, it \arm{remains unclear} whether or how inter-frame information extracted from motion or deformation can aid in \armj{identifying} objects}. }
{\yoshinew In addition, they all are based on popular convolutional generic still-image detectors~\cite{dai2016r,girshick2015fast,girshick2014rich,liu2016ssd,redmon2016yolo9000,ren2015faster} and it is not clear to what extent such generic object detectors, which are designed for and trained in dataset collected from the web, generalize to task-specific datasets~\cite{dollar2012pedestrian,hosang2015taking,zhang2016faster}. In the datasets for flying object detection that we use~\cite{trinh2016,rozantsev2017detecting}, the domain gap is especially large due to differences in the appearance of objects and backgrounds, \ari{as well as} the scale of the objects. Thus, we decided to use simpler region proposals and fine-tune our network as a region classifier for each dataset. }
%Thus, research questions should be 

\vspace{1mm}
\noindent {\bf Deep trackers} \hspace{1.5mm}
Recent studies have examined convnets and recurrent nets for tracking. Convnet-based trackers learn convolutional layers to acquire rich visual representations. Their localization strategies are diverse, including classification-based~\cite{nam2016learning}, similarity-learning-based~\cite{limulti}, regression-based~\cite{held2016learning}, and correlation-based~\cite{bertinetto2016fully,valmadre2017end} approaches. While classification of densely sampled patches~\cite{nam2016learning} has been the most accurate \arm{in} generic benchmarks, its computation is slow, and \seqsplit{regression-based}~\cite{held2016learning} and correlation-based ones~\cite{bertinetto2016fully,valmadre2017end} are used instead when the classification is to be done in real-time. Our network incorporates a correlation-based localization mechanism, having its performance enhanced by the representation shared by the detector.

Recurrent nets~\cite{werbos1988generalization,hochreiter1997long}, which can efficiently handle temporal structures in sequences, have been used for tracking~\cite{ning2016spatially,gordon2017re3,milan2017online,wang2017trajectory}. However, most utilize separate convolutional and recurrent layers and have a fully connected recurrent layer, which may lead to a loss of spatial information. \yoshiaapr{In particular, recurrent trackers have not performed as well as the best single-frame convolutional trackers in generic benchmarks.} \ari{One study used ConvLSTM with simulated robotic sensors for \rk{handling occlusions}} ~\cite{ondruska2016deep}.

\vspace{1mm}
\noindent {\bf Joint detection and tracking} \hspace{1mm} %短く
The relationship between object detection and tracking is a long-term problem in itself; before the advent of deep learning, it had only been explored with classical tools. \rk{In the track--learn--detection (TLD) framework~\cite{kalal2012tracking}, a trained} detector enables long-term tracking by re-initializing the trackers when objects disappear from view for a short period. Andriluka \etal \, uses a single-frame part-based detector and shallow unsupervised learning based on temporal consistency~\cite{andriluka2008people}. Tracking by associating detected bounding boxes \cite{huang2008robust} is another popular approach. However, in this framework, recovering undetected objects is challenging because tracking is more akin to post-processing following detection than to joint detection and tracking.

\vspace{1mm}
\noindent {\bf Motion feature learning} \hspace{1.5mm}
\rk{Motion feature learning (and hence recurrent nets) has been used for video classification~\cite{karpathy2014large} and action recognition~\cite{soomro2012ucf101}.} \yoshiapr{A number of studies have shown that \arm{LSTMs yield an improvement in accuracy}~\cite{wang2015action,weinzaepfel2015learning,donahue2015long}. For example, VideoLSTM~\cite{li2016videolstm} uses the idea of inter-frame correlation to recognize actions with attention. \rk{However, with \ari{action recognition} datasets, the networks may not fully utilize human motion features apart from appearances, backgrounds, and contexts ~\cite{he2016human}.}}

Optical flow~\cite{lucas1981iterative,horn1981determining,dosovitskiy2015flownet} is a pixel-level alternative to trackers~\cite{park2013exploring,gladh2016deep,zhu2016deep,zhu2017flow}. {\yoshinew Accurate flow estimation is, however, challenging in small flying object detection tasks because of the small apparent size of the targets and the large inter-frame disparity due to fast motions~\cite{rozantsev2017detecting}.} While we focus on high-level motion stabilization and motion-pattern learning via tracking, we believe flow-based low-level motion handling is orthogonal and complementary to our method, depending on the application area.

%-----------------------------------------------------------------------------------------------

\begin{figure}[t]
  \begin{center}
    \includegraphics[width=240pt]{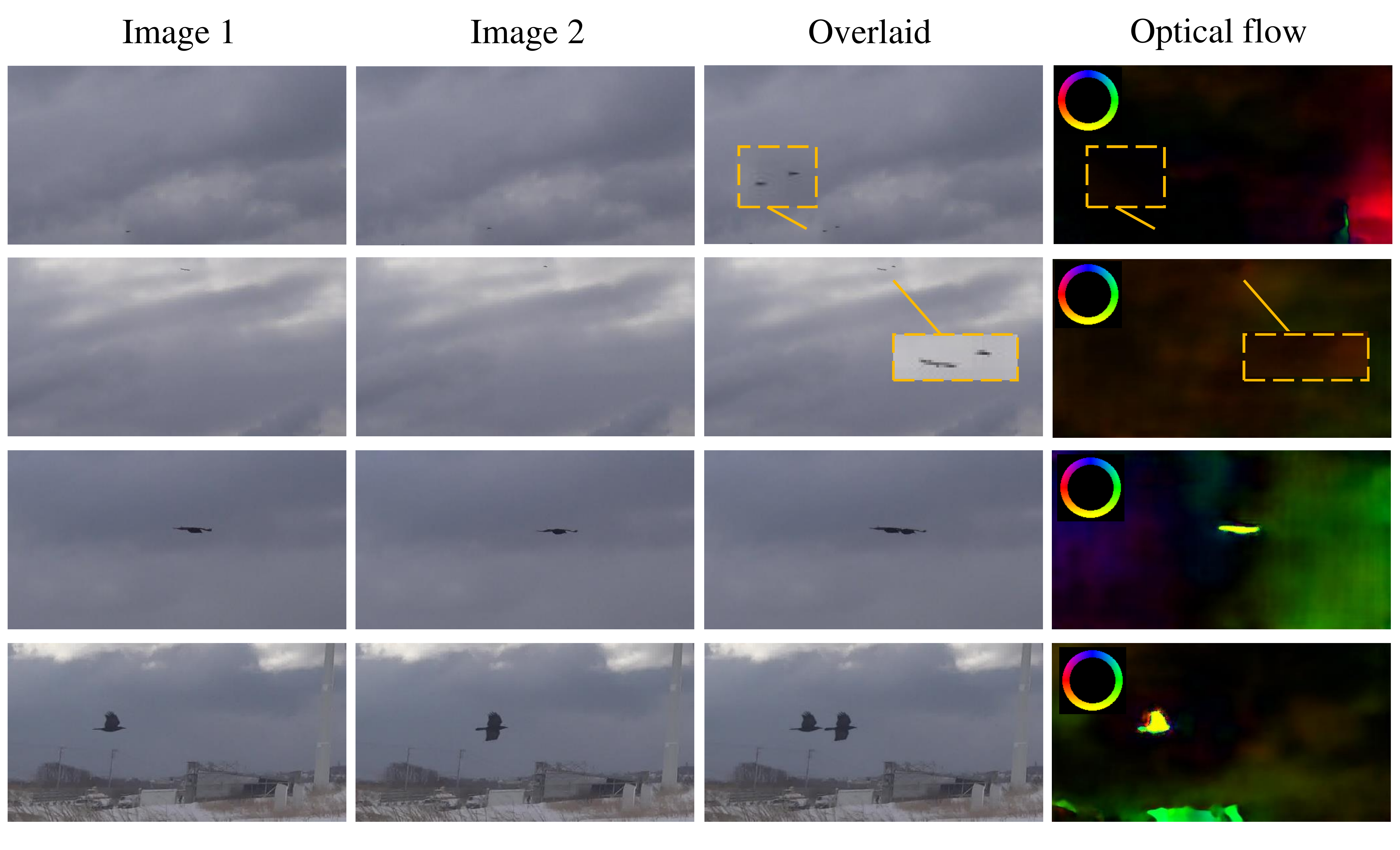}
  \end{center}
  \vspace{-4mm}
  \caption{Failures of optical-flow extraction in tiny bird videos. The lack of visual saliency of the foreground regions and high speeds of the birds prevented an accurate estimation of optical flow, on which many recent video-based recognition methods rely.}
  %\vspace{-5mm}
  \label{fig:flow}
\end{figure}

\begin{figure*}[t]%
  \begin{center}
    \includegraphics[width=480pt]{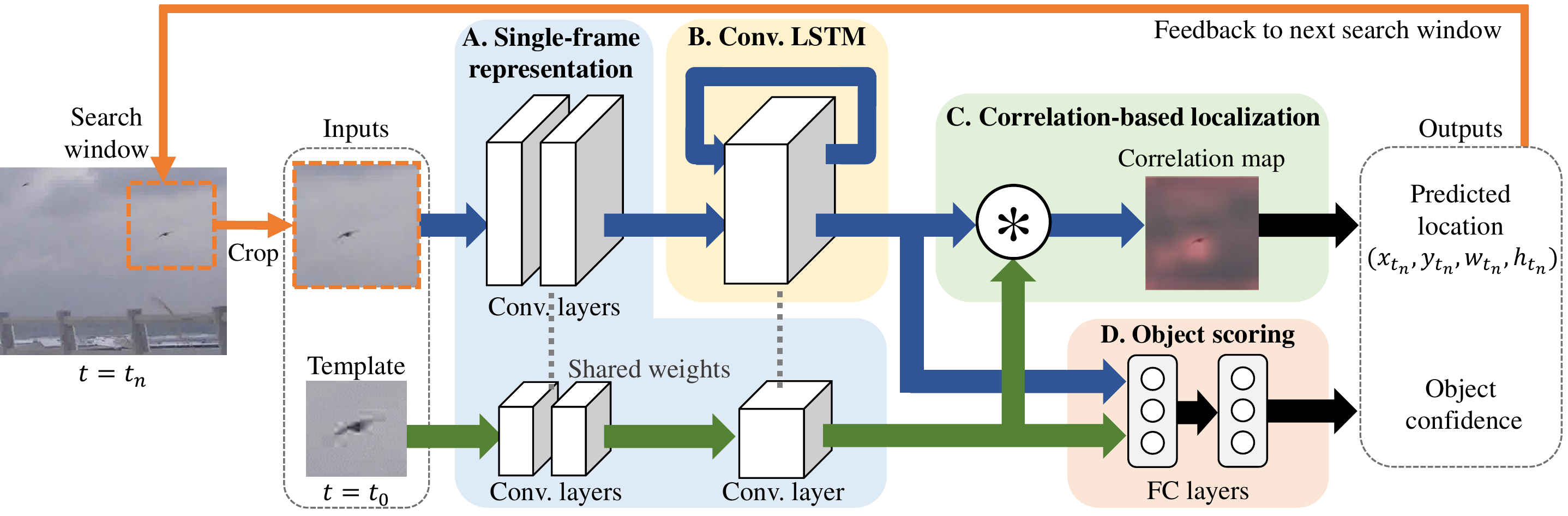}
  \end{center}
  \caption{Overview of the proposed network, called {\it Recurrent Correlation Network} (RCN). It consists of \yoshifinal{the four modules}: \yoshiaapr{Convolutional layers for single-frame representations (A), ConvLSTM layers for multi-frame representations (B), cross-correlation layers for localization (C), and fully-connected layers for object scoring (D)}. Green arrows show the information stream from templates \yoshiapr{(the proposals in the first frame at $t=t_0$)}, and blue arrows show that from search windows, \yoshijj{which keeps being updated by the tracking}.} 
  \label{fig:net}
\end{figure*}

\section{Recurrent Correlational Networks}
\yijcv{In this section, we describe the design of the {\it Recurrent Correlational Networks} (RCN) that especially aims to extract motion cues
from tiny flying objects via joint detection and tracking.
The key idea behind the RCN is that compensation of translation is essential for effective extraction of motion cues, due to the small size and relatively large movement of the objects.
This compensation is done
by simultaneous tracking with detection.
%Below, we first formalize our joint detection-and-tracking-based approach and later describe the architecture of the RCN.
Below, we first discuss the challenges of tiny flying object detection for further details, next formalize our joint detection-and-tracking-based approach, and later describe the architecture of the RCN.}

\subsection{\yoshijj{Challenges of tiny flying object detection in videos}}
Tiny birds are easier to find when they are moving---This intuition drove us to exploit motion cues 
for tiny flying object detection.
However, how to extract motion features that are powerful enough to differentiate tiny \armj{flying} objects is an open problem.
\yoshijj{The difficulty comes from the entangled nature of videos;
\armj{when} moving objects appear, \armj{the} temporal changes of \armj{the} video frames 
include \armj{the} objects' translations \armj{as well as} appearance changes.
While the \armj{translations} may \armj{not be so} useful, the \armj{appearance changes} include part-centric motions or deformations that often encode strong class-specific patterns, such as flapping of \armj{wings}, which
\armj{may} be \armj{very} useful \armj{for detection}.
When \armj{the translations are large}, such deformation patterns manifest \armj{themselves} as residuals only after the objects' translation is properly compensated.
Thus, to extract the discriminative patterns, we first need to disentangle \armj{the} translations and \armj{the} deformations by estimating the translations.}

\yoshij{\armj{However, estimating the translations of \yyijcv{tiny}, deforming, and fast-moving objects is a challenge in itself. }
The two major approaches \armj{used} to estimate motion vectors from video frames are optical flow and object tracking, but both approaches may fail with naive application to our challenging setting.
Optical flow \armj{refers to} dense motion descriptors that perform pixel-wise estimation of motion vectors.
They are hard to \armj{apply} to wide-area surveillance videos for \armj{two} reasons: First, computing of 
dense motion vectors is very time-consuming, especially when \armj{the} frame resolutions are large.
 Second, flows often exploit smoothness priors to resolve ambiguities and reduce noise.  
 Such priors are useful \armj{for improving the} accuracy of using optical flow on moderately large to large objects, but they may smooth out small objects and miss their motions \armj{entirely}. 
 Figure~\ref{fig:flow} shows examples of optical flow estimated by FlowNet-v2.0~\cite{ilg2016flownet}, confronting such difficulties. In the top two examples, small and non-salient birds were smoothed out. This seems to be due to the strong smoothness prior built in the optical flow method. In the bottom two examples, the optical flow noticed the birds, but the flow directions are not correct due to the disparity being relatively large for the object sizes. }
 
In contrast, tracking can be \armj{regarded} as a sparse, region-wise counterpart \armj{of optical flow that} can be efficiently applied to  small moving objects in high-resolution videos. 
\armj{Nevertheless}, robust tracking of small flying objects \armj{remains} challenging. Usually generic-object trackers are trained in a class-agnostic manner\armj{; they are trained in this way even when they are used on large amounts of video crawled from the Web.}
\armj{This lack of domain-specific knowledge} may make the trackers suboptimal in surveillance settings that handles a specific type of objects and scenes, and often causes many tracking failures in highly challenging scenes and \armj{with} \yoshijj{largely deforming} objects with low visibility.
To overcome these trackers' limitation, we design joint detection and tracking framework,
where the tracker, introduced to help the detector, itself is helped by the detector.

\subsection{\yoshijj{Joint detection and tracking formulation}}
\armj{Let us revisit the} formulations of conventional object detection and tracking, 
and extend them \armj{to} joint detection and tracking to give an overview of our framework.
Detection is a task to indicate objects in a frame by bounding boxes, and \armj{it} assigns 
detection confidence scores to the boxes.
\armj{A typical detector has} two \armj{stages}~\cite{girshick2014rich,ren2015faster}: 
\armj{the first stage} extracts object candidate boxes from the input image,
and \armj{the second} scores each of them by how likely it is \armj{to be} an object of interest.
\armj{A} single-frame-based detection algorithm is \armj{expressed} as follows:
\begin{eqnarray}
B = \{\bm{b}^t_0, \bm{b}^t_1, ... , \bm{b}^t_i, ..., \bm{b}^t_{N_t}\} &=& {\rm candidate}(\bm{I}_t), \nonumber \\
s^t_i &=& {\rm score}(\bm{b}^t_i, \bm{I}_t),
\end{eqnarray}
where $\bm{b}^t_i$ denotes the $i$-th bounding box in the $t$-th frame, $\bm{I}_t$ denotes the $t$-th frame, and $N_t$ \armj{is} the number of bounding boxes. 
$s^t_i$ is \armj{a} confidence score, \armj{where a higher value means a high probability of being an object.}
This framework of detection has no \armj{way} of exploiting temporal information.
\armj{A naive way of exploiting temporal information in multiple frames would be} as follows:
\begin{eqnarray}
B &=& {\rm candidate}(\bm{I}_t), \\
s^t_i &=& {\rm score}(\bm{b}^t_i, \{\bm{I}_t, \bm{I}_{t+1}, ..., , \bm{I}_{t+l}\}). \nonumber 
\end{eqnarray}
This allows the detector to access \armj{subsequent} frames to score a candidate box $\bm{b}^t_i$.
However, a problem \armj{with} this formulation is that \armj{it can not capture the} possible objects' movements:
if the object move from $\bm{b}^t_0$ in later frames, scoring it with reference to the original location $\bm{b}^t_0$ in $\bm{I}_{t+1}, \bm{I}_{t+2}, ..., , \bm{I}_{t+l}$ may be suboptimal 
because the box no longer pin-points the target.

\armj{Incorporating tracking in the detectors can solve the object movement problem.}
\armj{Like} detectors, trackers output bounding boxes 
but their difference is that trackers need to be initialized by the original location of a target object,
and then they keep indicating the object.
For simplicity, we \armj{will use} single-object tracking,
which can be denoted as follows:
\begin{eqnarray}
\bm{b}^t &=& {\rm track}(\bm{I}_t, \bm{z}_{t-1}), \nonumber \\
\bm{z}^t_i &=& {\rm update}(\bm{b}^t_i, \bm{I}_t, \bm{z}^{t - 1}), \\
\bm{z}^0 &=& {\rm initialize}(\bm{I}_0, \bm{b}^0), \nonumber 
\end{eqnarray}
where $\bm{z}^t$ denotes the state vector of the tracker 
that encodes temporal information. 
The simplest form of the state vector is
 templates cropped out from the initialization frame $I_0$, \armj{which} is used to perform localization of the target by matching the templates without updating~\cite{hager1998efficient,bertinetto2016fully}.
More sophisticated trackers have introduced discriminative optimization in \armj{the} initialization and updating to compute filters that separate the targets and backgrounds the best~\cite{bolme2010visual,danelljan2017discriminative}.
However, such trackers still need initialization by
the location of \armj{the} objects in the first frame, and
they do not encode semantic information of the tracked objects, 
in other words; they are not capable of detection.

To enable detectors to exploit multi-frame information,
we fuse the above detection and tracking framework into our joint detection and tracking, \armj{as} follows:
\begin{eqnarray}\label{eqns:jdt}
B = \{\bm{b}^0_0, \bm{b}^0_1, ... , \bm{b}^0_{N_t}\} &=& {\rm candidate}(\bm{I}_t), \nonumber \\
\bm{b}_i^t &=& {\rm track}(\bm{I}_t, \bm{z}^{t-1}_i), \nonumber \\
s^t_i &=& {\rm score}(\bm{b}^t_i, \bm{I}_t, \bm{z}^{t - 1}), \\
\bm{z}^t_i &=& {\rm update}(\bm{b}^t_i, \bm{I}_t, \bm{z}^{t - 1}), \nonumber \\
\bm{z}^0_i &=& {\rm initialize}(\bm{I}_0, \bm{b}^0_i). \nonumber
\end{eqnarray}
\armj{Unlike} the single-frame detectors, the confidence scores of objects depend on 
temporal states $\bm{z}^{t - 1}$ and \armj{the} updated locations of the objects $\bm{b}^t_i$. It is also different from trackers in that it outputs per-class confidence scores for detection, and it is initialized by region proposals in the first frame. 
The advantages of this joint-detection-tracking formulations are\armj{:} 
1) \armj{the detector} can exploit temporal contexts, including motions, in a natural manner \armj{through} fusion with \armj{the tracker}, 
and 2) by updating the bonding boxes of interest \armj{by using the tracker, the detector} can 
keep \armj{focused} on \armj{the} target objects in spite of their movement.

\vspace{-1mm}
\subsection{Architecture}
\yoshiapr{We \armj{designed} \arm{the} {\it Recurrent Correlational Network (RCN)} as shown in Fig.~\ref{fig:net}}, \armj{to enable joint detection and tracking with a deep convolutional architecture.} 
\arm{The network} consists of four modules: (A) convolutional layers, (B) ConvLSTM layers, (C) a cross-correlation layer, and (D) fully connected layers for object scoring. \armj{The} convolutional layers model single-frame appearances of target and non-target regions, including other objects and backgrounds. 
\armj{The} ConvLSTM layers encode temporal sequences of
single-frame appearances, and extract the discriminative motion patterns,
\yoshijj{which correspond to ${\rm update}$ in Eqns.~\ref{eqns:jdt}}.
%that are useful for classification.
%to be easily classified.}
\armj{The} cross-correlation layer convolves the \rk{representation} of the template \armj{with} that of \armj{the} search windows in subsequent frames, and generates correlation maps that are useful for localizing the targets,
\yoshijj{which is corresponding to ${\rm track}$ in Eqns.~\ref{eqns:jdt}}.
Finally, the confidence scores of the objects are calculated with fully-connected layers based on the multi-frame representation,
\yoshijj{which corresponds to ${\rm score}$ in Eqns.~\ref{eqns:jdt}}.
\yoshifinal{
The network is supervised by the detection loss,
and the tracking gives locational feedback for \armj{the} region of interest in \armj{the} next frames during training and testing.
%is also aided via the shared multi-frame representation via ConvLSTM.
}

{\yoshinew
Our detection pipeline is based on region
proposal and classification of the proposal,
as in region-based CNNs~\cite{girshick2014rich}.
The main difference is that our joint detection and tracking network simultaneously track the given proposals in the following frames, and the results
of the tracking are reflected in the classification scores that are used as \armj{the} detectors' confidence scores. 
%The inputs to the network are external object proposal boxes and video frames. 
}

\vspace{1.5mm}
\noindent {\bf Convolutional LSTM} \hspace{1.5mm}
In our framework, the ConvLSTM module~\cite{xingjian2015convolutional} is used for motion feature extraction (Fig.~\ref{fig:net}~B). It is a convolutional counterpart of LSTM~\cite{hochreiter1997long}. It replaces \armj{the} inner products in the LSTM with convolutions, \yoshifinal{\armj{which are} more suitable for motion learning, since the network is more sensitive to local spatio-temporal patterns rather than \armj{to} global patterns.}
It works as a sequence-to-sequence predictor; specifically, it
takes \armj{a} series $(x_1, x_2, x_3, ..., x_t)$ of single-frame representations whose length is $t$ as input, and outputs a merged single representation $h_t$, 
at each timestep $t = 1, 2, 3, ..., L$.

For the sake of completeness, we show the formulation of ConvLSTM below. 
\begin{eqnarray} \label{eqn:lstm} 
i_t &=& \sigma(w_{xi} * x_t + w_{hi} * h_{t - 1} +b_i) \nonumber \\ 
f_t &=& \sigma(w_{xf} * x_t + w_{hf} * h_{t - 1} + b_f) \nonumber \\
c_t &=& f_t \circ c_{t - 1} + i_t \circ \mathrm{tanh}(w_{xc}  * x_t + w_{hc}  \circ h_{t - 1} + b_c) \nonumber \\
o_t &=& \sigma(w_{xo} * x_t + w_{ho} * h_{t - 1} + b_o) \nonumber \\
h_t &=& o_t \circ \mathrm{tanh}(c_t).
\vspace{4cm}
\end{eqnarray}
Here, $x_t$ and $h_t$ \armj{respectively} denote the input and output of the layer at timestep $t$, respectively. The states of the memory cells are denoted by $c_t$. $i_t$, $f_t$, and $o_t$ and are called gates, which work for selective memorization. `$\circ$' denotes the Hadamard product.
\yoshijj{In our framework, ($h_t$, $c_t$) composes the context vector $\bf{z}_t$ in Eqns.~\ref{eqns:jdt}.}
ConvLSTM is also well suited to exploit the spatial correlation for joint tracking, since its output representations are in 2D.
%To exploit the spatial correlation, the aggregated multi-frame representations need to be in 2D. The recently developed ConvLSTM is well suited to fulfilling the requirement of the recurrent module . 

While ConvLSTM is effective at video processing, it inherits the complexity of LSTM. The gated recurrent unit (GRU) is a simpler alternative to LSTM that has fewer gates, and it is empirically easier to train on some datasets~\cite{chung2015gated}. A convolutional version of the GRU (ConvGRU)~\cite{siam2016convolutional} is as follows: 
\begin{eqnarray}
z_t &=& \sigma(w_{xz} * x_t + w_{hz} * h_{t - 1} + b_z) \nonumber \\ 
r_t &=& \sigma(w_{xr} * x_t + w_{hr} * h_{t - 1} + b_r)  \\ 
h_t &=& z_t \circ h_{t-1} + (1 - z_t) \nonumber \\ 
	& & \circ \hspace{1mm} \mathrm{tanh}(w_{xh} * x_t + w_{hh} * (r_t \circ h_{t - 1}) + b_h) .\nonumber 
\vspace{4cm}
\end{eqnarray}
ConvGRU has only two gates, \yoshiapr{namely an update gate $z_t$ and reset gate $r_t$}, while ConvLSTM has three.
ConvGRU can also be incorporated into our pipeline; later we provide an empirical comparison between ConvLSTM and ConvGRU.

\begin{figure}[t]%
  \begin{center}
    \includegraphics[width=235pt]{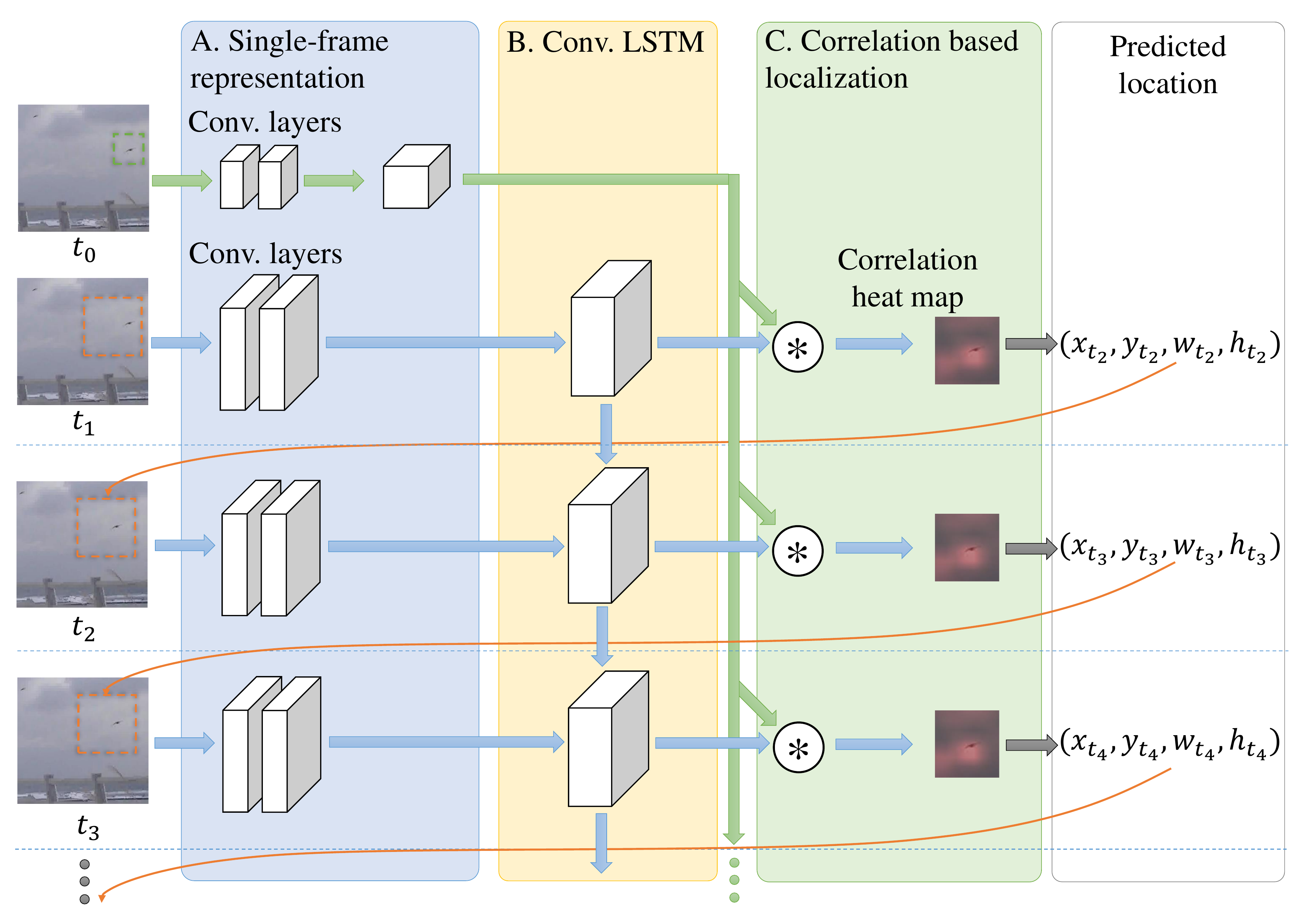}
  \end{center}
  \caption{Temporal expansion of the proposed network. The joint tracking is incorporated as part of the feedback in the recurrent cycle. 
  \yoshifinal{This feedback provides stabilized observation of moving objects, 
  while learning from deformation is difficult without stabilization \yyijcv{when the translation is large}.}} 
  \label{fig:tex}
\end{figure}

\vspace{1mm}
\noindent {\bf Correlation-based localization} \hspace{1.5mm} 
The correlation part (Fig.~\ref{fig:net}~C) aims to stabilize \armj{a} moving object's appearance by tracking.
The localization results are fed back to the next
input, as shown in Fig.~\ref{fig:tex}. 
This feedback allows ConvLSTM to learn deformations and pose changes apart from \armj{the} translation, while \armj{the} local motion patterns
are invisible \armj{because of} translation without stabilization.

Cross-correlation is an operation that relates two inputs and outputs a correlation map that indicates how similar \arm{a} patch in an image is to another. It is expressed as
\begin{equation}
\vspace{-1mm}
C(\bm{p}) = \bm{f} * \bm{h} =  \sum_{\bm{q}} \bm{f}(\bm{p} + \bm{q})  \cdot \bm{h}(\bm{q}). \\ \vspace{-1mm}
\end{equation}
where $\bm{f}$ and $\bm{h}$ denote the multi-dimensional feature representations of the search window and template, respectively. $\bm{p}$ is for every pixel's coordinates in the domain of $\bm{f}$, and $\bm{q}$ is for the same but in the domain of $\bm{h}$. \armj{The} two-dimensional (2D) correlation between \armj{the} target patch and \armj{the} search window is equivalent to densely comparing the target patch with all possible patches within the search window.  The inner product is used here as \armj{the} similarity measure. 

In the context of convolutional neural networks, the cross-correlation layers can be considered to be differentiable layers without learnable parameters; namely, a cross-correlation layer is a variant of the usual convolutional one whose kernels are substituted by the output of another layer. Cross-correlation layers are bilinear with respect to two inputs, and thus are differentiable. The computed correlation maps are used to localize the target by 
\vspace{-0mm}
\begin{equation}
\bm{p}_{target} = \mathrm{argmax}_{\bm{p}} C(\bm{p}) 
\vspace{-0mm}
\end{equation} 

\vspace{1mm}
\noindent {\bf Single-frame representation} \hspace{1.5mm} 
A multi-layer convolutional representation is inevitable in natural image recognition, although the original ConvLSTM ~\cite{xingjian2015convolutional} did not use non-recurrent convolutional layers in radar-based tasks. Following recent tandem CNN-LSTM models for video recognition~\cite{donahue2015long}, we insert non-recurrent convolutional layers before the ConvLSTM layers (Fig.~\ref{fig:net}~A). 
Arbitrary convolutional architectures can be
incorporated and we should choose the proper
ones for each dataset.
We experimentally tested two different structures of varying depth.  

We need to extract an equivalent representation from the object template for the search windows. For this, we use ConvLSTM, in which the recurrent connection is severed. Specifically, we force the forget gates to be zero and  enter zero vectors instead of the previous hidden states. This layer is equivalent to a convolutional layer with $tanh$ and sigmoid gates. It shares weights with $w_{xc}$ in Eq. \ref{eqn:lstm}.

\vspace{1mm}
\noindent {\bf Search window strategy} \hspace{1.5mm} 
In object tracking, as the \armj{physical} speed of the target objects is physically limited, limiting the area of the search windows, where the correlations are computed, is a natural way to reduce computational costs. We place windows the centers of which are at the previous locations of \arm{the} objects; \yoshiapr{\armj{the windows have} a radius $R = \alpha \max(W, H)$, where $W$ and $H$} are  the width and height of the bounding box of the candidate object. We then compute the correlation map for \armj{the} windows around each candidate object. 
%This determination of the search windows is part of a feedback mechanism in the recurrent part, which is shown in Fig.~\ref{fig:tex}. 
We empirically set the size of the search windows to $\alpha = 1.0$.
The representation extracted from the search windows is also fed to the object scoring part of the network, which yields large field-of-view features and provides contextual information for detection. %We set the size of the search windows to $\alpha = 2$ based on experiments.

\vspace{1mm}
\noindent {\bf Object scoring} 
\hspace{1.5mm} 
For object detection, the tracked candidates need to be scored according to likeness. We use fully connected (FC) layers for this purpose (Fig.~\ref{fig:net}~D). We feed both the representations from the templates (green lines in Fig.~\ref{fig:net}) and the search windows (blue lines in Fig.~\ref{fig:net}) into the FC layers by concatenation. We use two FC layers, where the number of dimensions in the hidden vector \armj{is} 1,000. 

We feed the output of each timestep of ConvLSTM into the FC layers and average the scores. In theory, the representation of the final timestep after feeding \armj{in} the last frame of the sequence should provide the maximum information.  However, we found that the average scores are more robust in case of tracking failures or the disappearance of targets.

\subsection{Details}
\vspace{1mm}
\noindent {\bf Multi-target tracking} 
\hspace{1.5mm} 
In surveillance \armj{situations}, many object candidates \armj{may} appear in each frame, and we need to track them simultaneously for joint detection and tracking.
However, correlation-based tracking \armj{was}
originally designed for single-object tracking
and \armj{its} extension to \armj{the} multiple object tracking is non-trivial.
We extended \armj{it in} the following manner.
First, \armj{we contatenate} $N$ \armj{cropped} regions and templates into a four-dimensional array
\armj{in the} shape of $(N, 3, W, H)$ or $(N, 3, w, h)$,
where $W$, $H$, $w$, and $h$ are the widths or heights of the search windows and templates. Then we 
compute the forward pass of the network
and acquired $(N, 1, W, H)$ correlation maps with in a single forward computation.
\armj{The implementation reuses} \armj{the} convolution layers 
with \armj{a} small modification \armj{so that is} can inherit
the efficiency of \armj{a} heavily optimized GPU computation.

\paragraph{Inference algorithm} 
\begin{algorithm}[t]
\caption{\yoshij{RCN inference algorithm.}}
\label{alg:inference}
\begin{algorithmic}
\REQUIRE Video frames $I_1, I_2, ..., I_T$, object candidates $B = \{\bf{b}_1, \bf{b}_2, ..., \bf{b}_N\}$, trained RCN network $\texttt{RCN}$.
\ENSURE The candidates' object-likelihood scores $s_1, s_2, ..., s_N$
\STATE Initialize RCN's hidden states $\bf{h}_{i,1} \leftarrow \bf{0}$ for $i=1, ..., N$. 
\STATE Initialize objects' locations $\bf{b}_{i,1} \leftarrow \bf{b}_i$ for $i=1, ..., N$. 
\STATE $\bf{x}_i \leftarrow \texttt{crop}(I_1, \bf{b}_i)$ for $i=1, ..., N$.//Crop templates from the initial frame.

\FOR{t = 2 \texttt{to} T}
    \FOR{i = 1 \texttt{to} N}
        \STATE $\bf{w}_{i,t} \leftarrow \texttt{expand}(\bf{b}_{i,t-1}$) // Expand the object boxes and use them as search windows.
        \STATE $\bf{z}_{i,t} \leftarrow \texttt{crop}(I_t, \bf{w}_{i,t})$ 
        \STATE $s_{i,t}, b_{i,t}, h_{i,t} = \texttt{RCN}(\bf{x}_i, \bf{z}_{i,t}, \bf{h}_{i,t-1})$ //Compute the forward path
    \ENDFOR
\ENDFOR
\STATE $s_{i} = \texttt{average}(s_{i,1}, ..., s_{i,T})$
\end{algorithmic}
\end{algorithm}

\yoshij{Our total inference algorithm is iterations \armj{of} two steps:
a feed-forward computation of RCN and 
a re-cropping of the updated search windows from the next time-step frame.
\armj{The} pseudo code is
 \armj{shown} in Algorithm~\ref{alg:inference}.}

\vspace{1mm}
\noindent {\bf Training} 
\hspace{1.5mm} 
Our network is trainable with ordinary gradient-based optimizers in an end-to-end manner, because all layers are differentiable. 
%However, we use scheduling techniques to ensure fast convergence. 
\yoshifinal{We separately train \armj{the} convolutional parts
and \armj{the} ConvLSTM to ensure fast convergence and avoid overfitting.} 
\armj{First, we} initialize single-frame-based convnets by \armj{using the} pre-trained weights in the ILSVRC2012-CLS dataset, the popular and large generic image dataset. We then fine-tune single-frame convnets \armj{on} the target datasets (birds, drones, and pedestrians) without ConvLSTM. Finally, we add \armj{a} convolutional LSTM, correlation layer, and FC layers to the networks and fine-tune them again.  For optimization, we use the SGD solver of Caffe~\cite{jia2014caffe}. \armj{In the experiments reported below}, the total number of iterations was 40,000, and the batch size was five. The original learning rate was 0.01, and \armj{it} was \yoshiapr{reduced by a factor of 0.1} per 10,000 iterations. The loss was the usual sigmoid cross-entropy \yoshifinal{for detection.} 
%and the tracking is not explicitly supervised}.
We freeze the weights in the pre-trained convolutional layers after connecting \armj{them} to the convolutional LSTM to avoid overfitting. 

\armj{For} training \armj{the} ConvLSTM, we use pre-computed trajectories predicted by a
single-frame convolutional tracker, which consists of the final convolutional layers of the pre-trained single-frame convnet and a correlation layer. \armj{The trajectories} are slightly inaccurate but \armj{are} similar to those of our final network. 
\armj{We} store \armj{the} cropped search windows in the disk during \armj{the} training for efficiency, to reduce disk \armj{accesses} by avoiding the re-cropping of the regions of interest out of the 4K-resolution frames during \armj{the} training.
During the \armj{test} phase, the network observes trajectories \rk{estimated} by itself, which are different from the ground truths \armj{used} in the training phase.  
This training scheme is often referred to as teacher forcing~\cite{williams1989learning}.  Negative samples also need trajectories in training, but we do not have their ground truth trajectories because only the positives are annotated in the detection datasets.  

\paragraph{Trajectory smoothing}
\yoshij{\armj{Although} our network can robustly track \armj{small} objects, we also found that post hoc smoothing of the trajectories further improves
the localization accuracy \armj{when targets disappear temporarily}. 
For \armj{this} purpose, we adopted Kalman filter with \armj{a} constant-velocity dynamic model. In \armj{the tracking experiments}, we additionally \armj{computed} the tracking accuracy \armj{when} this smoothing \armj{was used}.}

%\begin{comment}

\section{\yoshij{Dataset construction}}
While flying-object surveillance is practically important, the number and diversity of publicly available datasets are limited. Thus, we constructed a video bird dataset to enable large-scale evaluations of flying-object detection and tracking. Here, we describe the construction method and properties of the dataset. 

\begin{figure}[t]
  \begin{center}
       \includegraphics[width=235pt]{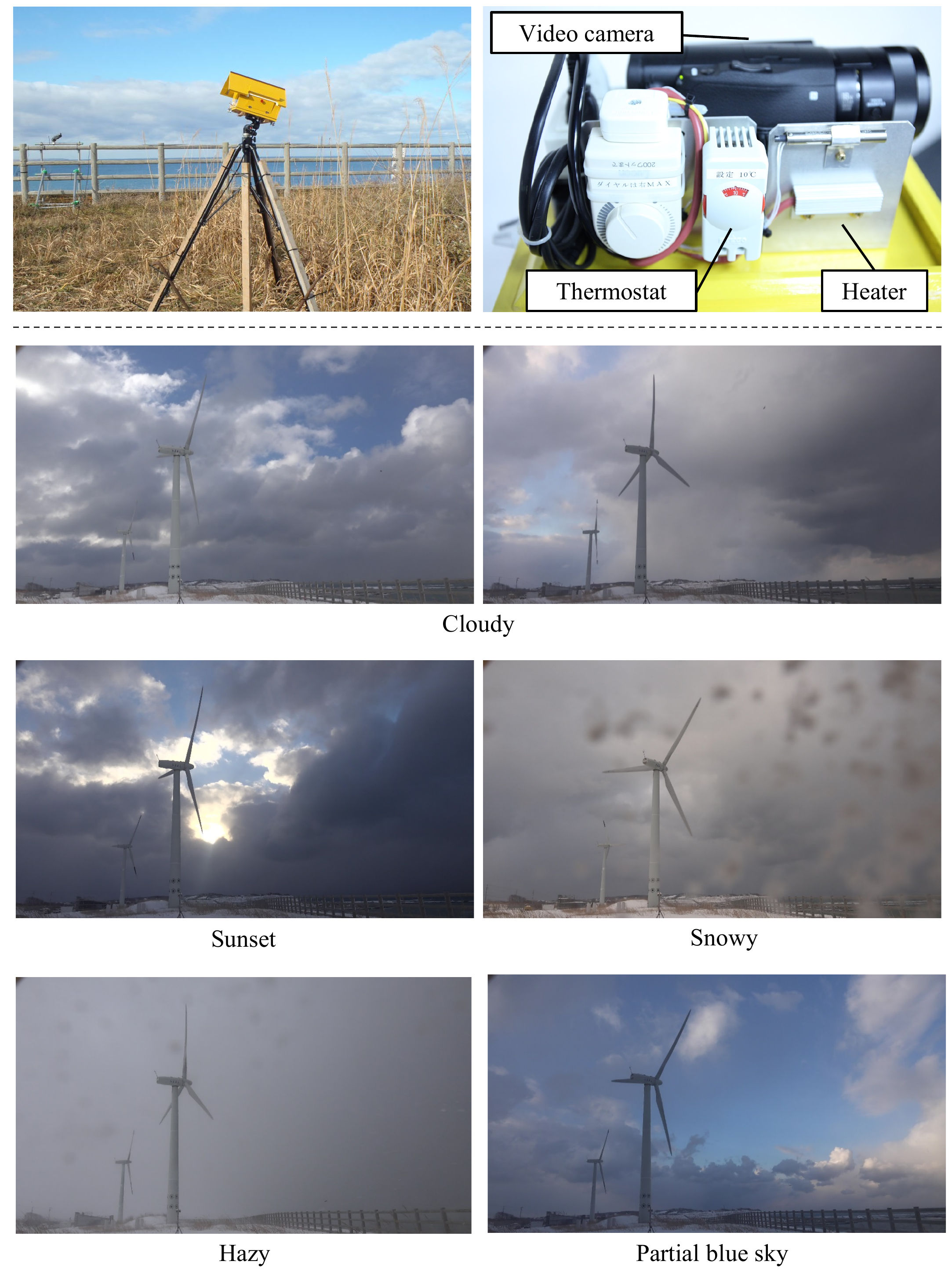}
  \end{center}
  \vspace{-4mm}
  \caption{Setup for capturing video and examples of captured videos. The videos include challenging variations of weather and illumination.}
  \vspace{-6mm}
  \label{fig:videosetup}
\end{figure}

\paragraph{Video recording}
We set up a fixed-point video camera at a wind farm. We selected the location in connection with a project to monitor endangered birds' collisions with the turbines. We recorded the video in the daytime (8:00--16:00) for 14 days. Among the recorded videos, we selected 3 days' worth of videos with relatively frequent appearances of birds and annotated them. The videos were in 4K UHDTV ($3840 \times 2160$) resolution and stored in MP4 format, which made the file size 128GB per day. Despite the high resolution, compression noise was visible on the fast moving objects in the images. Figure~\ref{fig:videosetup} shows our recording setup together with heating equipment to remove snow.

\paragraph{Statistics}

\begin{figure}[]
  \begin{center}
    \includegraphics[width=230pt]{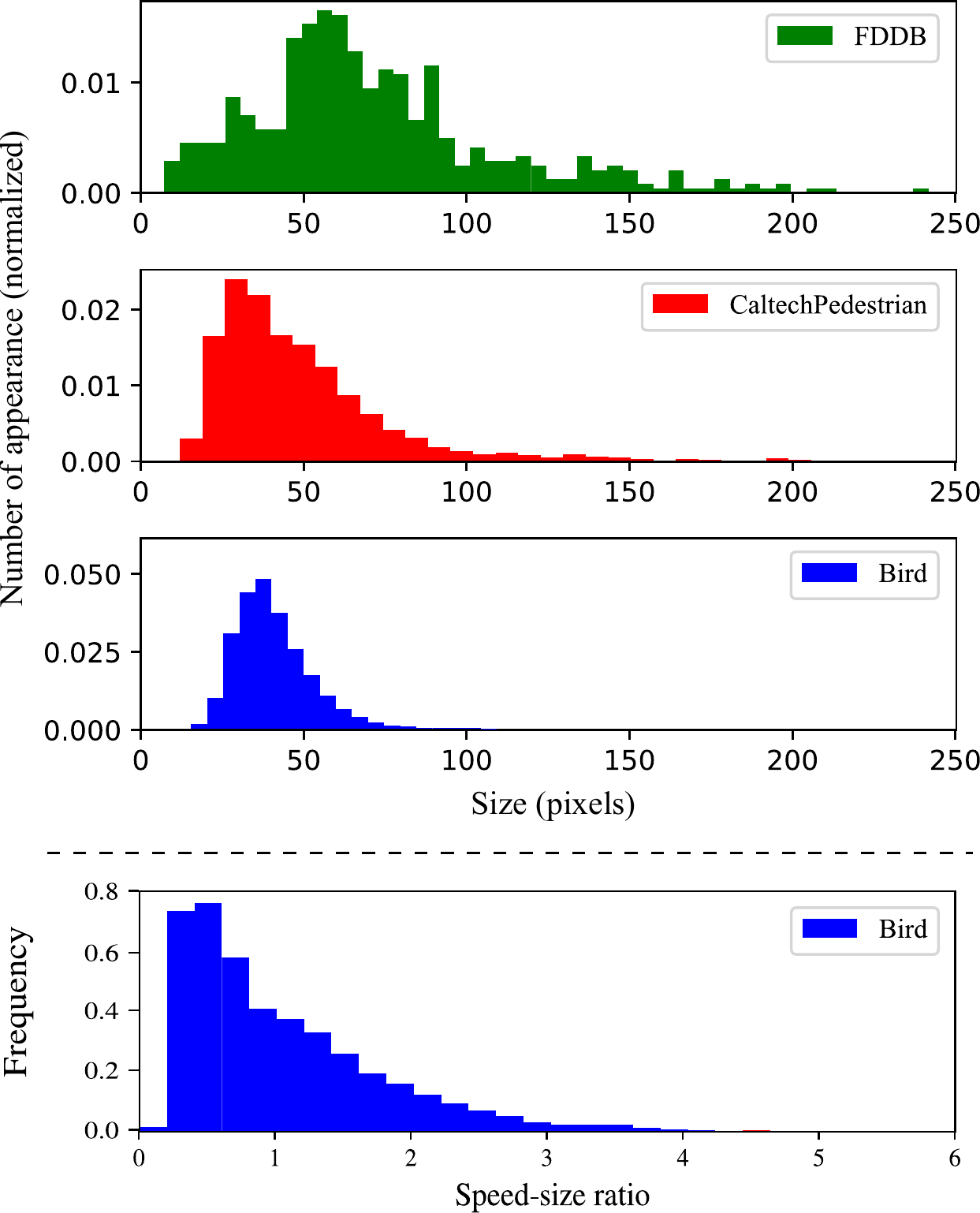}
    %\vspace{-0mm}
    %\includegraphics[width=250pt]{speed.pdf}
  \end{center}
  \vspace{-4mm}
  \caption{The statistics of our bird dataset and comparisons to existing detection datasets~\cite{fddbTech,dollar2012pedestrian}. Upper: The distribution of annotated objects' size. Lower: The distribution of annotated birds' moving speed's ratio to the bird sizes. \yyijcv{In the bird dataset, small objects appear more often despite of the 4K resolution of full frames, and their movements are often large for their sizes.}}
  %\vspace{-5mm}
  \label{fig:dists}
\end{figure}

Figure~\ref{fig:dists} shows the distribution of bird sizes and speeds. The bird sizes were measured by the longer sides of their bounding boxes, their widths in most cases. The mode of the size distribution is 25 pixels. This is smaller than the mode of most existing detection datasets, including datasets of pedestrians~\cite{dollar2012pedestrian}, faces \cite{fddbTech}, and generic objects~\cite{everingham2010pascal}. Furthermore, birds fly quickly for their small size. About the half of the birds moved more than their boxes' longer side between consecutive frames (Fig.~\ref{fig:dists} lower). This means the optical flows and trackers must be robust to large disparities.


%-----------------------------------------------------------------------------------------------
\section{Experiments}
\begin{figure*}[]
  \begin{center}
    \hspace{-4mm}
    \includegraphics[width=500pt]{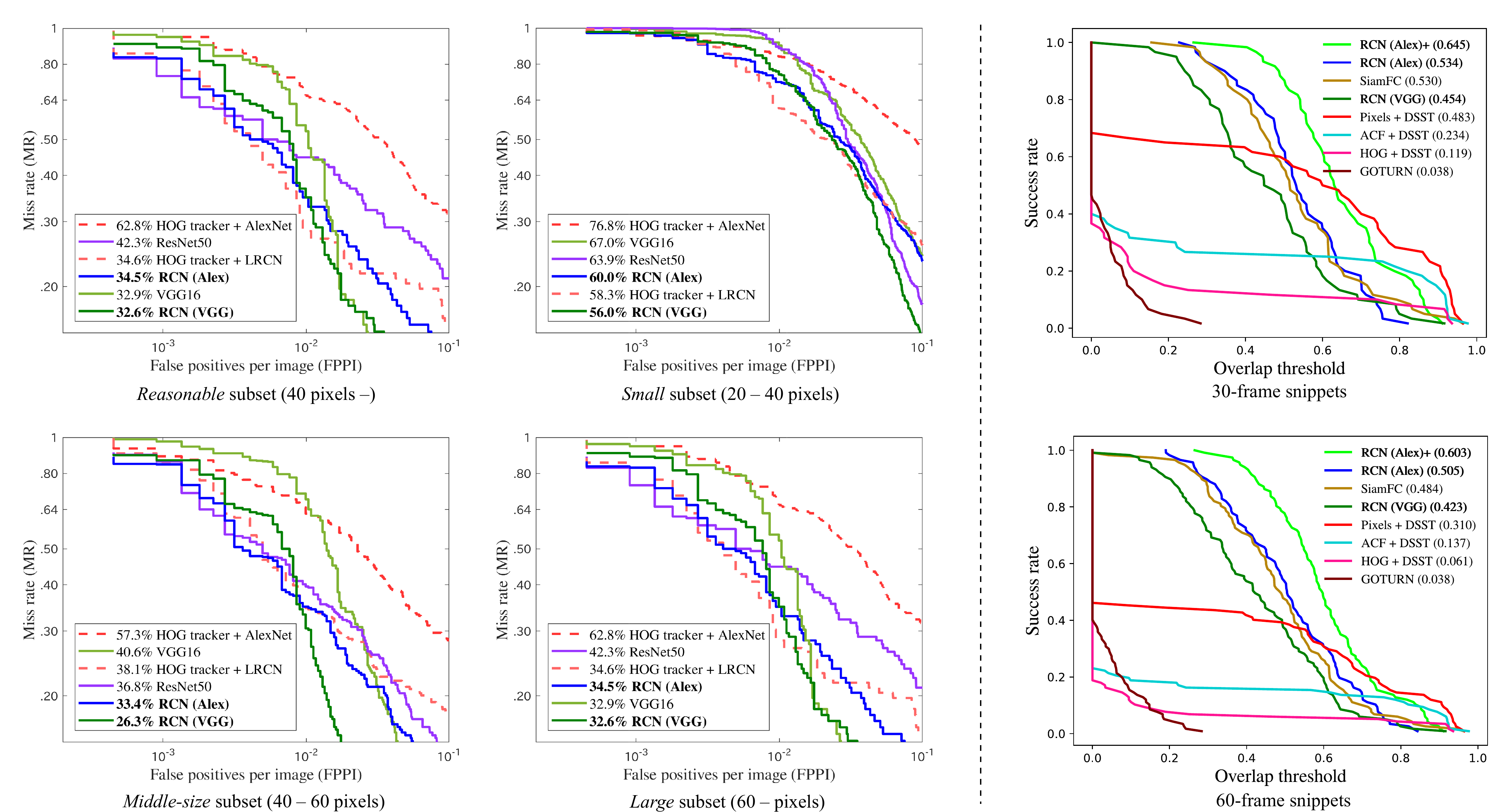}
  \end{center}
  \vspace{-3mm}
  \caption{Left: Bird detection results. The lower left is better. \textbf{Our RCN (VGG)} outperformed \armj{all} the other methods with deeper convolutional layers, and \textbf{our RCN (Alex)} outperformed the previous method with the same convolutional layer depth on three subsets. The subsets are \armj{distinguished} by the sizes of \armj{the} birds in the images. Right: Bird tracking results. The upper right is better. \armj{Our methood} outperformed DSST trackers with various handcrafted features and ImageNet-pretrained deep trackers.}
  \label{fig:rocs}
\end{figure*}

The main purpose of the experiments was to investigate the performance gain owing \yoshiapr{to the learned motion patterns with joint tracking in small object detection tasks}. We also investigated the tracking performance of our method and compared it with that of \yoshiaapr{trackers with a variety of features
\ari{as well as} convolutional trackers}. 

\subsection{Experimental settings}
%\vspace{-2mm}
\yoshijj{To evaluate our method's performance for small flying object, we first used \armj{the bird video dataset} described in Section 4}.
We also tested our method on a UAV dataset~\cite{rozantsev2017detecting} to see whether  it \armj{could} be applied to other flying objects. This dataset consists of 20 sequences of hand-captured videos. It \armj{has} approximately 8,000 bounding boxes of flying UAVs. All the UAVs are multi-copters. We followed the training/testing split provided by the authors of ~\cite{rozantsev2017detecting}. \yoshiapr{The properties of the dataset are summarized in Table ~\ref{tab:datasets}.}

\yoshij{Additionally, we applied our method to a more general computer-vision task, i.e., pedestrian detection and tracking. While recent pedestrian detectors exploit only appearance-based features, pedestrians in images are often barely visible or appear blurred, and motion patterns such as gait are expected to aid recognition. For this experiment, we used the Caltech Pedestrian Detection Benchmark (CPD), one of the largest datasets focusing on pedestrians.} 

\begin{table}[tb]
	\begin{center}
    \caption{Statistics of the datasets used in the experiments.}
    \includegraphics[width=\linewidth]{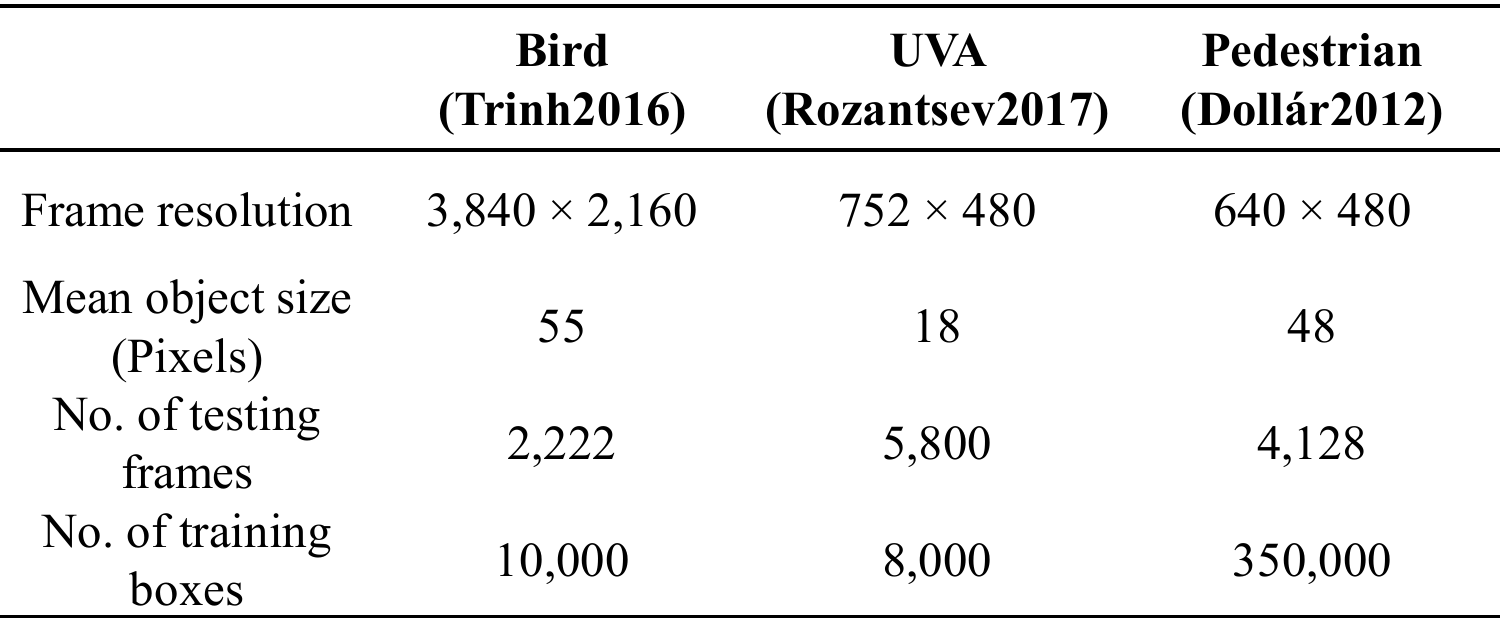}
    \vspace{-8mm}
  \label{tab:datasets}
  \end{center}
\end{table}

\paragraph{Evaluation metric}
To evaluate detection performance,  we used the number of false positives per image (FPPI) and the log average miss rate (MR). 
These metrics were based on single-image detection; {\em i.e.}, they were calculated only on given test frames that were sampled discretely. Detection was performed on the given test frames and, for our method, tracking of { \yoshinew all} candidates was conducted in some of the subsequent frames. We used the toolkit \yoshij{originally} provided for the Caltech Pedestrian Detection Benchmark~\cite{dollar2012pedestrian} to calculate the scores and plot the curves in Fig.~\ref{fig:rocs}.

We also tested \armj{the} tracking accuracy separately from \armj{the} detection on the bird detection dataset. 
{\yoshinew We fed the ground-truth bounding boxes in the first frames to our network and other trackers, aiming to evaluate our \armj{network} as a tracker.}
We conducted one-path evaluation (OPE), tracking by using ground truth bounding boxes given only in the first frame of the snippets without \yoshiaapr{re-initialization, re-detection, or trajectory fusion}. To \armj{avoid evaluating the trackers on} very short trajectories, we selected ground-truth trajectories longer than 90 frames (three seconds at 30 fps) from the annotation of the bird dataset.  We plotted success rates versus overlap thresholds. The curves in the right of Fig.~\ref{fig:rocs} show the proportion of the estimated bounding boxes whose overlaps with the ground truths were higher than the thresholds. 

\paragraph{Object proposals}
We used a different strategy for each dataset to generate object proposals for pre-processing. In the bird dataset, we extracted the moving object by background subtraction~\cite{zivkovic2004improved}. The extracted regions were provided \armj{with} the dataset; therefore, we could compare the networks fairly, regardless of the hyperparameters or the detailed tuning of the background subtraction. \armj{On} the UAV dataset, we used the HOG3D-based sliding window detector provided by the authors of ~\cite{rozantsev2017detecting}. 
\yoshij{\armj{On the} pedestrian datasets,
we use a region proposal net \armj{(RPN)} that were tuned for pedestrian detection~\cite{zhang2016faster} without any modification.}

\paragraph{Compared methods}
\armj{In the results described below,} {\it RCN~(Alex)} and {\it RCN~(VGG)} \armj{denote} two implementations of the
proposed method using the convolutional
layers from AlexNet~\cite{krizhevsky2012imagenet} and VGG16Net~\cite{Simonyan15}.
{\it HOG tracker+AlexNet} and {\it HOG tracker+LRCN} are baselines for the bird dataset provided by ~\cite{trinh2016}.  The former is a combination of the histograms of oriented gradients (HOG)-based \cite{dalal2005histograms} discriminative scale-space tracker \\(DSST \cite{danelljan2014accurate,danelljan2017discriminative}) and convnets that classify the tracked candidates into positives and negatives. The latter is a combination of DSST and the CNN-LSTM tandem model~\cite{donahue2015long}.
\armj{In the experiments, they used five frames following the test frames. For a fair comparison, our method used the same number of frames in the detection evaluation.}.
In addition, we fine-tuned VGG16Net \cite{Simonyan15} and ResNet50 \cite{he2016deep} as still-image-based baselines.

\armj{To evaluate} the tracking performance, we included other combinations of the DSST and hand-crafted features for further analysis. {\it HOG+DSST} is the original version in ~\cite{danelljan2014accurate}. {\it ACF+DSST} replaces the classical HOG with more discriminative aggregated channel features~\cite{dollar2014fast}. The \armj{aggregated channel feature (ACF)} is 
similar to HOG,
%\rv{also gradient-based like HOG}, 
but is more powerful because of the additional gradient magnitude and LUV channels for orientation histograms. {\it Pixel+DSST} is a simplified version that uses RGB values of raw pixels instead of gradient-based features. 
\yoshiapr{ We also included ImageNet-pretrained convolutional trackers, namely, correlation-based SiamFC~\cite{bertinetto2016fully} and regression-based GOTURN~\cite{held2016learning}. They are based on the convolutional architecture \arm{of} AlexNet.} 

\begin{figure*}[]
  \begin{center}
    \includegraphics[width=485pt]{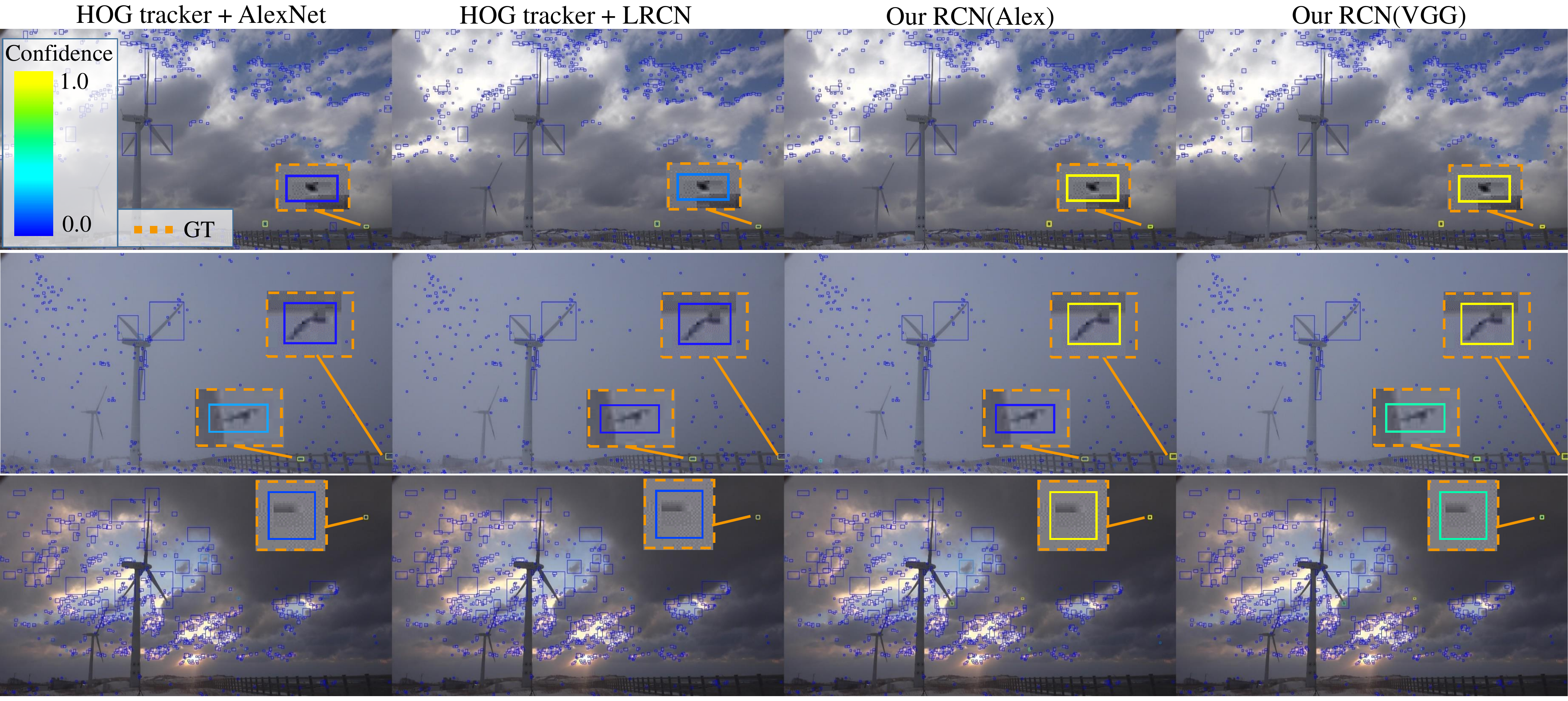}
  \end{center}
  \vspace{-6mm}
  \caption{{\yoshinew Example frames of results of detection on the bird dataset~\cite{trinh2016}. \yoshi{The dotted yellow boxes show ground truths, enlarged to avoid overlapping and keep them visible. The confidence scores of vague birds are increased and \armj{those} of non-bird regions are decreased by our RCN detector. The contrast was modified for visibility in the zoomed-up samples.}}}
  \label{fig:bird_result_examples}
  \begin{center}
    \includegraphics[width=485pt]{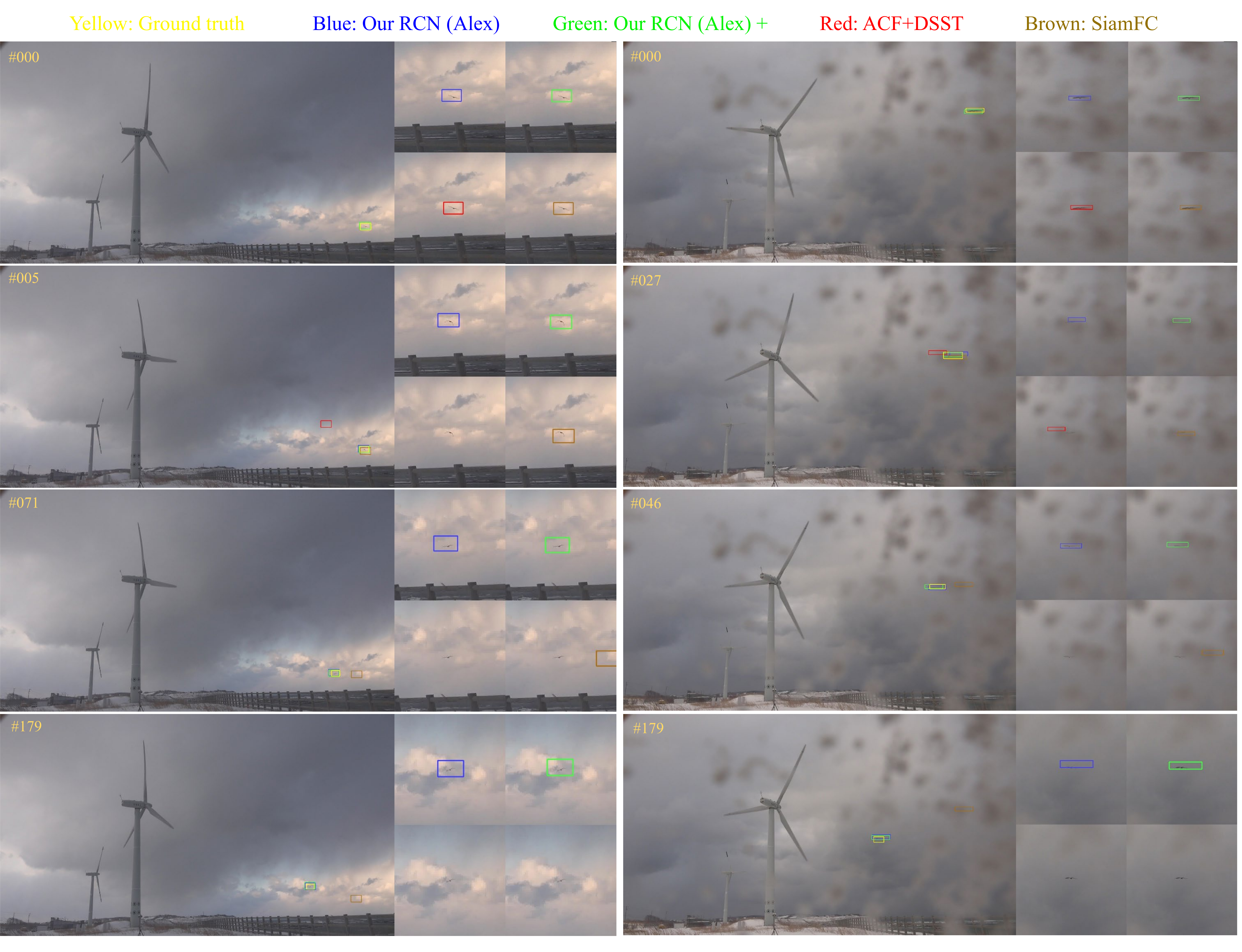}
  \end{center}
  \vspace{-6mm}
  \caption{\yoshij{Examples of bird tracking results. Our trackers RCN (Alex) (blue) and RCN (Alex)+ (green) track the small birds more robustly, whereas generic-object trackers with hand-crafted features (DSST, red) and deeply learned features (SiamFC, brown) tend to miss the targets in low visibility frames. RCN (Alex)+ performed a more accurate localization than RCN (Alex) did, owing to the trajectory smoothing. More examples are shown in the supplementary video.}}
  \label{fig:trk_examples}
\end{figure*}
\subsection{Results}
\paragraph{Bird Detection and Tracking Results}
The results of detection on the bird dataset are shown in Fig.~\ref{fig:rocs}. The curves are for four subsets of the test set, which consists of birds of different sizes, namely {\it reasonable} (over 40 pixels square),  {\it small} (smaller than 40 pixels square),  {\it mid-sized} (40--60 pixels square), and {\it large} (over 60 pixels square).

On all subsets, the proposed method, {\it RCN~(VGG)} showed the smallest average miss rate (MR) of the tested detectors. 
The improvements \armj{in comparison with the previous best published method {\it HOG tracker+LRCN}} were -10.3 percentage points on {\it Reasonable}, -2.3 percentage points on {\it Small}, 
-14.4 on {\it Mid-sized}, and -0.9 percentage points on  {\it Large}.

A comparison of {\it HOG tracker+LRCN} and 
{\it RCN~(Alex)} is also important,
because \armj{they} share the same convolutional architecture.
\armj{Here,} {\it RCN~(Alex)} performed better on all of the subset except {\it Small}.
The margins \armj{were} -3.5 percentage points on {\it Reasonable}, -4.7 percentage points on  {\it Mid-sized} subset, and -0.1 percentage points on {\it Large} subset.
Examples of the test frames and results are shown in Fig.~\ref{fig:bird_result_examples} (more examples are in the supplementary material).

A comparison of {\it RCN~(Alex)} and {\it RCN~(VGG)} provides an interesting insight. {\it RCN~(Alex)} \armj{was} more robust against smaller FPPI values in spite of the lower average performance than that of {\it RCN~(VGG)}.
{\it RCN (Alex)} \armj{had} a smaller MR than {\it RCN~(VGG)} when the FPPI was lower than $10^{-2}$. A possible reason is that
a deeper network is less generalizable because \armj{it has} many parameters; thus, it may miss-classify new negatives more often in the test set than the shallower one.
%\rv{the effect of the network depth, and the effect of larger sizes of input images in {\it RCN (Alex)}.}

The results of tracking on the bird dataset are shown in the right of Fig. \ref{fig:rocs}.
%\shaodi{please fix it}. 
We found that gradient-based features were inefficient on this dataset. HOG-based DSST missed the target even when tracking for 30 frames (this is already longer than what was used in ~\cite{trinh2016} for detection). We supposed that this failure was due to the way the HOG normalizes the gradients, which might render it over-sensitive to low-contrast but complex background patterns, like clouds. We found that replacing HOG with ACF and utilizing gradient magnitudes and LUV values benefited the DSST on the bird dataset. However, the simpler pixel-DSST outperformed the ACF-DSST by a large margin. 

The trajectories provided by our network were more robust than all of the DSST variations tested. This shows that representations learned through detection tasks also work better in tracking than hand-crafted gradient features do. It also worth noting that our trajectories were less accurate than those obtained through the feature-based DSSTs when they did not miss the target. When bounding-box overlaps larger than 0.6 were needed, the success rates were smaller than those of the DSSTs for both 30- and 60-frame tracking. This is because our network used a correlation involving a pooled representation, the resolution of which was 32 times smaller than that of the original images. \yoshiapr{In addition, RCN (Alex) outperformed two convnet-based trackers (GOTURN and SiamFC). \yoshij{RCN (Alex)+, the combination of ours with the Kalman filter, further boosted tracking performance.} Examples of tracking results are presented in the supplementary material.}

\begin{figure}[!t]
  \begin{center}
    \vspace{-0mm}
    \hspace{-0mm}
    \includegraphics[width=260pt]{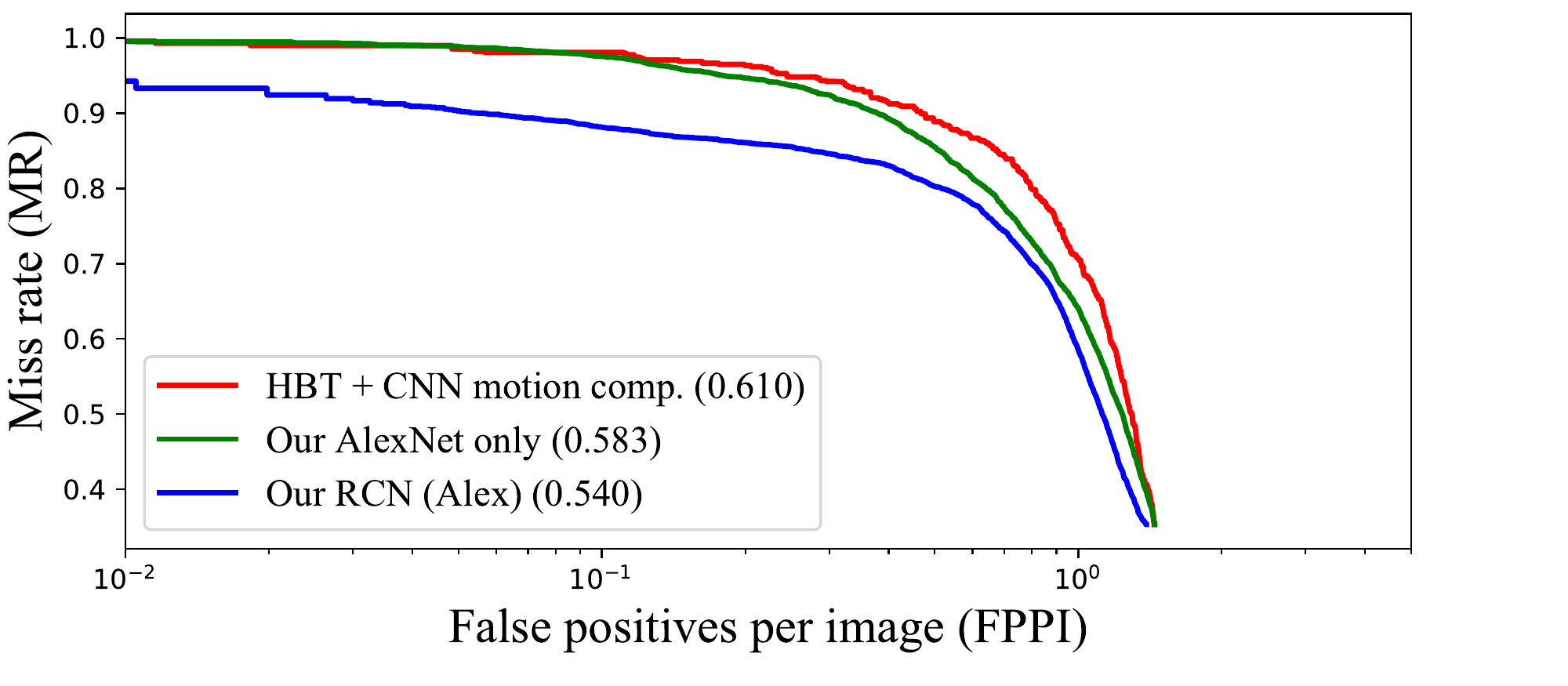}
  \end{center}
  \vspace{-4mm}
  \caption{Detection results on the UAV dataset~\cite{rozantsev2017detecting}. RCN performed the best.}
  \vspace{-6mm}
  \label{fig:roc_uav}
\end{figure}

\begin{figure}[!t]
  \begin{center}
    \includegraphics[width=240pt]{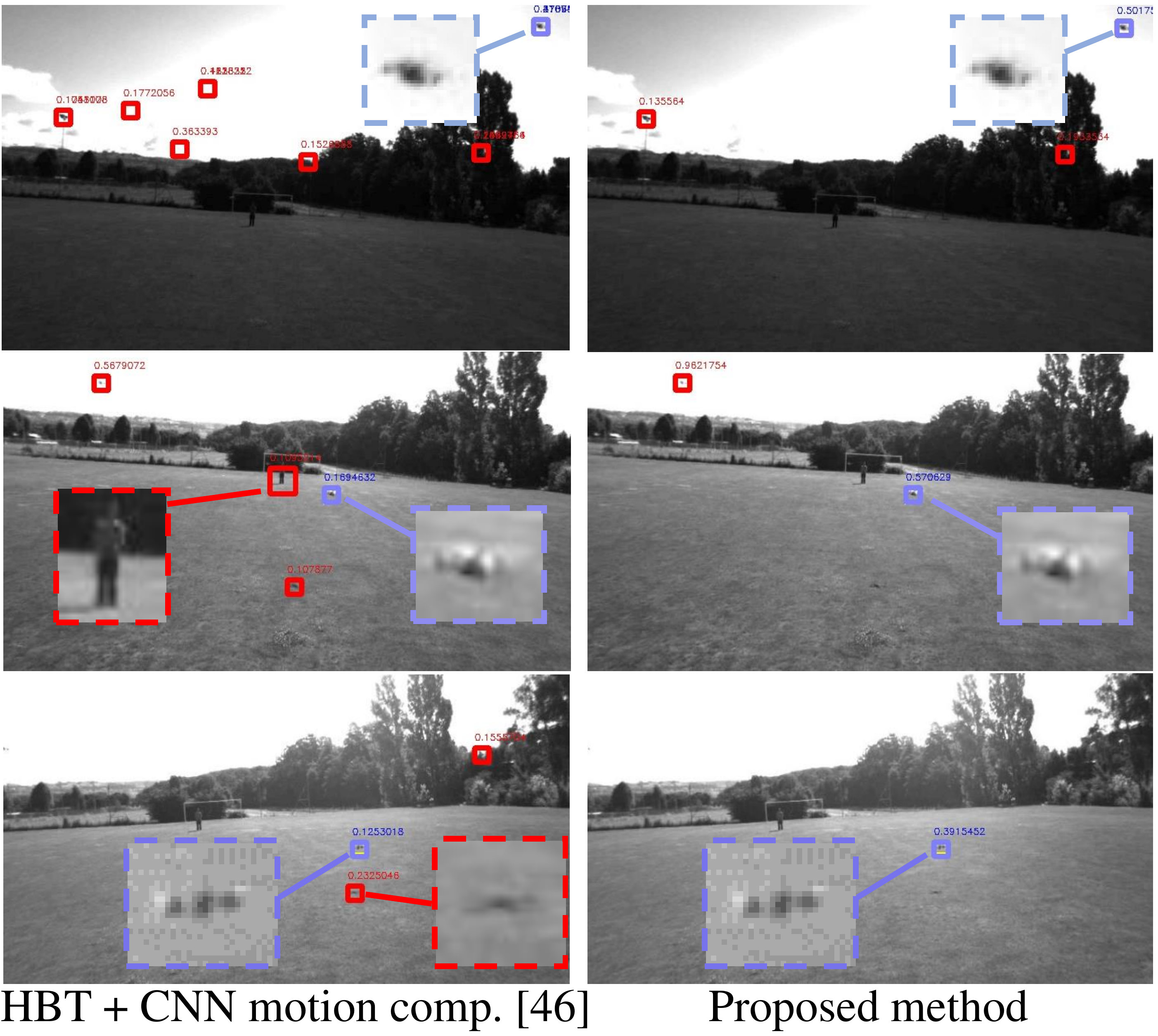}
  \end{center}
  \vspace{-2mm}
  \caption{Sample frames of detection results on the UAV dataset~\cite{rozantsev2017detecting}. The blue boxes
  show correct detections and the red ones show misdetections. \yoshi{Our method \armj{made} fewer misdetections when the detectors thresholds were set to give roughly the same MR.}}
  \vspace{-0mm}
  \label{fig:drone_samples}
\end{figure}

\paragraph{Drone Detection Results}
The ROC curves of the drone detection are shown in Fig.~\ref{fig:roc_uav}. \armj{The results are for} a shallower AlexNet-based version of RCN, \armj{because there was not much training data}. \armj{The results for} AlexNet after single-frame pre-training without LSTM or tracking ({\it Our AlexNet only}) slightly outperformed the baseline in ~\cite{rozantsev2017detecting} without  multi-frame information, because {\it Our AlexNet only} was deeper and larger, and had been pre-trained in ImageNet.  
The pre-training on the ImageNet classification turned out to be useful even for small, grayscale UAV detection. The ConvLSTM and joint tracking consistently improved detection performance (-4.3 percentage points). However, the performance gain was smaller than that on the bird dataset. The reason seemed to be that the amount motion information in the UAV dataset was limited because the objects were rigid, in contrast to the articulated deformation of birds. Examples of the results are shown in Fig.~\ref{fig:drone_samples}.
%\vspace{-4mm}

\begin{table}[tb]
  \begin{center}
    \caption{MR \armj{on} Caltech Pedestrian with the new annotation. Ours achieved competitive detection performance compared to the state-of-the-art pedestrian detectors.}
    \label{tab:respeddet}
    \scriptsize
    \begin{tabular}{|l|c|c|c|}\hline
          & Method & MR & Year  \\ \hline \hline
          Existing  & ACF &  27.6 & PAMI14 \\
          models    & LDCF &  23.7 & NIPS14 \\
          & CCF    & 22.2 & ICCV15 \\
          & Checkerboard &  18.5  &  CVPR15 \\
          & DeepPart    &  10.64  & ICCV15 \\
          & TLL-TFF         &  10.3 & ECCV18 \\
          & MS-CNN          & 9.50  & ECCV16 \\
          & FasterRCNN   &  8.70 & ICCV17 \\
          & CompACTD    &  7.56 & ICCV15 \\
          & UDN+         &  8.47  & PAMI18 \\
          & PCN       &  6.29  & BMVC17 \\ 
          & SDS-RCNN  &  \bf{5.57} & ICCV17 \\ \hline
         Our & RPN & 10.22 & --  \\
         models & VGG    & 8.70 & -- \\
          & RCN $l=1$   & 9.22 & -- \\
          & RCN $l=5$   & \bf{7.83} & --  \\ \hline
         \vspace{0mm}\hspace{-3mm} \begin{tabular}{l} Combina- \\ torial \end{tabular} & CCF+CF         & 19.5  & ICCV15 \\
         \begin{tabular}{c}  \hspace{-3mm}\vspace{-2mm} models \\ \  \end{tabular}  & RPN+BF    &  7.32 & ECCV16 \\
           & HyperLearner    & \bf{5.30}   & ICCV17 \\ \hline
    \end{tabular}
    \vspace{-5mm}
  \end{center}
\end{table}

\begin{table}[!t]
	\begin{center}
     \caption{Ablation study: performance differences as a result of varying models and parameters. MR represents the log-average miss rate \armj{on} the {\it reasonable} subset of the bird dataset, and diff. represents its difference from the baseline. $k$ denotes the kernel size of the ConvLSTM.}
     \label{tab:ablation}
    \begin{scriptsize}
  \begin{tabular}{|cl|c||c|c|} \hline
         & & Network config. & MR & diff. \\ \hline \hline
       &  \hspace{-6mm} RCN (Alex)& & & \\
       		  & k = 3 & A + B + C + D & {\bf 0.336}&   0  \\ %\hline
    		  & k = 1 & A + B + C + D & 0.346  &   + 0.010  \\ %\hline
              & k = 5 & A + B + C + D & 0.347  &   + 0.011 \\ \hline
              %& w/o TC&           0.355  &   + 0.19  \\ \hline
              & \hspace{-6mm} RCN (VGG)& &  & \\
  	          & k = 3 & A + B + C + D & {\bf 0.268} &   0 \\ %\hline
              & ConvGRU k = 3 & A + B + C + D & 0.271  &   + 0.003  \\ 
              & w/o tracking& A + B + D & 0.321  &   + 0.053  \\ 
              & w/o ConvLSTM& A + C + D & 0.344  &   + 0.076  \\
    		  & Single frame& A + D & 0.332  &   + 0.064  \\ \hline
              
  \end{tabular}
  \end{scriptsize}
  \end{center}
\end{table}

\paragraph{Pedestrian Detection Results}
\yoshij{The results shown in Table~\ref{tab:respeddet} summarizes the MR of our and other recent methods on a {\it Reasonable} subset of the CPD. Note that the {\it our models} part of the table includes results for our model and its ablations, while the {\it combinatorial models} part includes results for combinations of existing models. In particular, the {\it our methods} section of the table compares RCN (VGG), i.e., {\it RCN $l=5$}, and its ablations. 
Our region proposal network (RPN) re-trained on CPD did not perform very well (MR 10.22). 
{\it RCN $l=5$}'s MR was 2.4 \% lower than the RPN, meaning that it missed 25\% fewer pedestrians.
Moreover, against a simple {\it VGG}, which rescored the region proposals by using the vanilla VGG16 net, and {\it RCN $l=1$}, which was our network but only using single frames,{\it RCN $l=5$} outperformed both. These results show that our method was effective at pedestrian detection.}

\yoshij{Our method performed comparably to some of the state-of-the-art pedestrian-specific detectors. It outperformed recent detectors, including the vanilla Faster RCNN~\cite{mao2017can}, ComPACTD~\cite{Cai15}, and UDN+~\cite{ouyang2018jointly}. It also outperformed the most recent detector that utilizes multi-frame information and a ConvLSTM (TTL-TFA~\cite{song2018small}, MR 10.22).}

\yoshij{The methods that outperformed ours utilized techniques specialized to pedestrian detection, for example, manually designed part models (PCN~\cite{wang2018pcn}), joint segmentation and detection (SDS-RCNN~\cite{brazil2017illuminating}), or a combination of hand-crafted and deep features (HyperLearner \cite{mao2017can}). Our method does not exploit ad hoc techniques tailored especially for pedestrian detection and is conceptually much simpler. Thus, we conclude that exploiting motion information via joint detection and tracking will be useful in a wide range of applications.}

\begin{table}[tb]
	\begin{center}
     \caption{Relationship between detection performance and numbers of inference timesteps.}
     \label{tab:timevar}
	%\vspace{-2mm}
    \scriptsize
  \begin{tabular}{|c|ccccc|} \hline
        \#Time steps at test & 1 & 3 & 5 & 7 & 9  \\ \hline \hline
         Training with $l=5$ & 0.305 & 0.274 & {\bf 0.268}  & 0.274 & -- \\
        Training with $l=8$ & 0.386 & 0.355 & {\bf 0.345} & {\bf 0.345} & 0.354 \\ \hline

  \end{tabular}
  \vspace{-3mm}
  \end{center}
  %\vspace{-5mm}
\end{table}

\begin{figure}[t]
  \begin{center}
    \includegraphics[width=240pt]{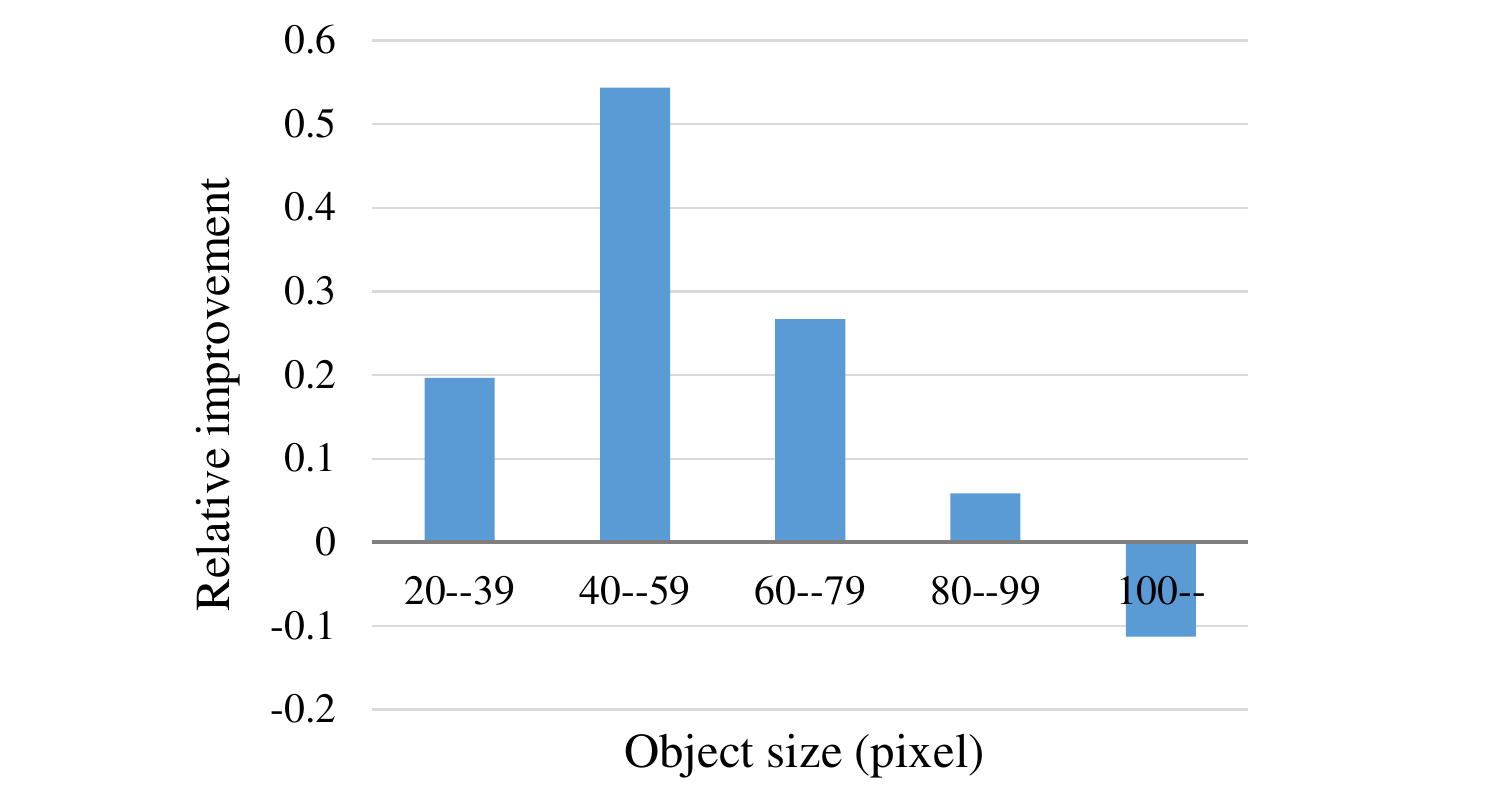}
  \end{center}
  \vspace{-4mm}
  \caption{\yijcv{Relative improvement in MR on different scales by introducing motion cues. The improvements for small objects (20 -- 79 pixels) are significant, which indicates the importance of motion when detecting small objects.}} %ToDo: refer in the main text
  \vspace{-5mm}
  \label{fig:scale_improvements}
\end{figure}

\subsection{Hyperparameters and ablation}
Here, to provide further insights into our model, we report the performance for different settings of the \ari{networks} and hyperparameters (Table \ref{tab:ablation}). Here, {\it Network config.} indicates which modules in Fig. \ref{fig:net} are active. All of the results were obtained from the {\it reasonable} subset of the bird dataset. 

\paragraph{Kernel size in ConvLSTM}
\yoshijj{First, we investigated the effect of different kernel sizes in the ConvLSTM. The kernel size is a hyperparameter that controls the receptive field of a memory cell. A ConvLSTM with too small kernels may not be able to handle spatiotemporal information, while one with too large kernels may be inefficient and cause overfitting. In our architecture, $k = 3$ was the best (MR 0.336); larger or smaller kernels has a slightly adverse effect on performance (+0.011 and + 0.010 MR). We used $k = 3$ in all of the later ablations, by default.}

\paragraph{Recurrent net variants}
\yoshijj{Second, we checked the effect of varying the recurrent architecture, specifically by replacing the ConvLSTM with a ConvGRU. The performance of the ConvGRU was only slightly worse than that of ConvLSTM (+0.003 MR), possibly because the input was pre-processed by convolutional layers and the burden on the recurrent part was smaller.}

\paragraph{Without tracking}
\yoshijj{ Even without the external stabilization by tracking, the ConvLSTM itself may have the ability to learn patterns from moving objects to some extent. Here, we investigated how much joint detection-tracking benefits the ConvLSTM in spatiotemporal learning. The ConvLSTM without tracking surely improved detection performance to some extent (-0.011 MR from the single-frame model), but it did not match that of the full model (+0.053 MR). This shows that stabilization by tracking is needed in order to fully exploit motion information in our framework.}

\paragraph{Without recurrence}
\yoshijj{ Fourth, we removed the recurrent part and averaged the confidence scores over time, to see the importance of the recurrent part. Without the recurrent part, the network could not learn spatiotemporal patterns; it only could learn spatial patterns and temporally average them. The averaging still may benefit detection by smoothing out hard-to-recognize frames, and if our network can learn motion patterns, it should be outperform the simple smoothing. In fact, the model without the recurrent part (w/o ConvLSTM) performed much worse than the full model (+0.076 MR).}

%\paragraph{Single frame}
%{Finally, we used the \yoshifinal{network} as a single-frame detector.}

{\yoshinew Overall, we found that a lack of stabilization, recurrent parts, or multi-frame cues led to critical degradations in performance; these results demonstrate the effectiveness of our network design.}

\paragraph{Number of timesteps}
\yoshij{Table~\ref{tab:timevar} summarizes the relationship between the number of time steps in testing and MR. Not surprisingly, the models performed the best when the numbers of inference time steps in training and testing were equal, because it gives the best match between the training and testing temporal feature distributions. We additionally trained a model with longer training snippets ($l=8$). Training with $l=8$ required a larger video memory, so we reduced the training batch size to half of $l=5$; this resulted in worse convergence. However, it consistently performed the best when the number of time steps in the test was equal to the number of time steps in the training.}

\begin{figure*}[t]
  \begin{center}
    \includegraphics[width=490pt]{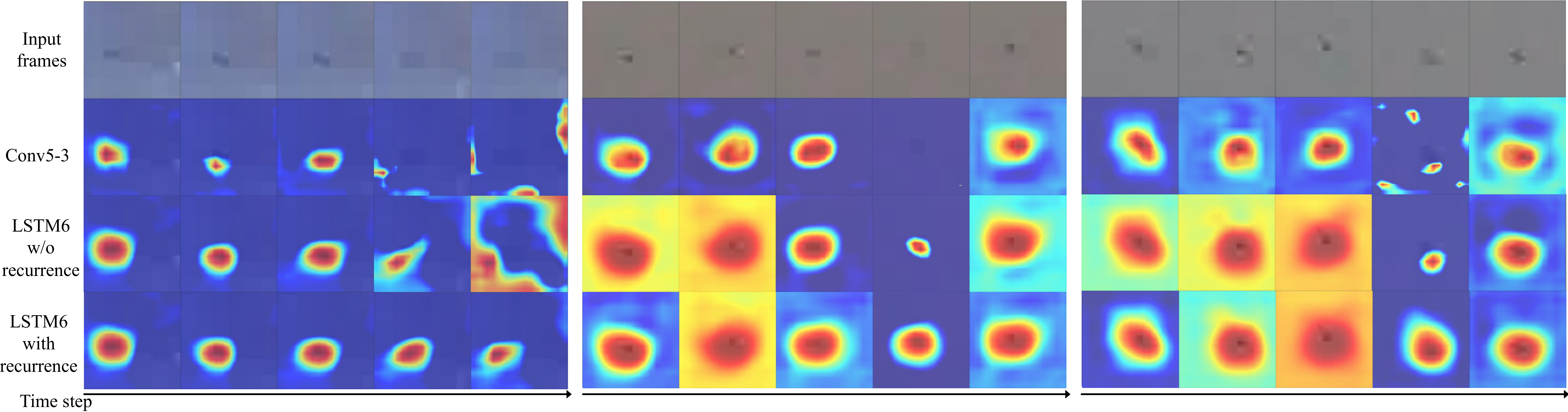}
  \end{center}
  \vspace{-4mm}
  \caption{Visualization of the multi-frame feature activation using Grad-CAM~\cite{selvaraju2017grad}.}
  \vspace{-5mm}
  \label{fig:gradcam}
\end{figure*}

\paragraph{Object size vs. improvement by motion cues}
\yijcv{Figure \ref{fig:scale_improvements} plots relative improvement in bird detection MR by exploiting motion cues, i.e., 
our full RCN (VGG)'s performance gain against, the single-frame baseline.
The models are the same as in Table~\ref{tab:ablation}.
The improvement for small objects (20$--$79 pixels) are as large as 20\% to 50\%, and 
this result supports our hypothesis that {\it motion cues are crucial in tiny object detection.}}

\subsection{Visualization}
\yoshij{Finally, we visualized the effects of motions on the learned multi-frame representations by using the Grad-CAM~\cite{selvaraju2017grad} method. GradCAM is useful for visualizing the contributions from each region in the input images on the per-class feature activation.}

\yoshij{In our framework, the recurrent connections of ConvLSTM are needed to extract motion cues that differentiate class-specific motion patterns. To understand their importance, we compared class activations from three layers: Conv5-3, ConvLSTM6 without recurrence, and ConvLSTM6 with recurrence. Conv5-3, the final convolutional layer (corresponding to Fig.~\ref{fig:net} A), is the most natural choice to see the single-frame activations. In addition, we visualized the single-frame activation of ConvLSTM6, the recurrent part (corresponding to Fig.~\ref{fig:net} B) by removing the recurrent connection. This enables a comparison of the same module with and without the recurrent connections and this is useful for understanding their role.}

\yoshij{Figure~\ref{fig:gradcam} shows the Grad-CAM mapping results. In time steps where the visual input was poor, single-frame activations in Conv5-3 and ConvLSTM6 w/o recurrence often became weak as can be seen in the 4th frame in (a), the 4th frame in (b), or the 4th and 5th frames in (c). In contrast, ConvLSTM6 with recurrence could attend to the non-salient inputs in such frames. This suggests that the relationships between sequential frames that were learned by the recurrent connections guided the attention of the network. }

\section{Discussion}

\paragraph{Relationship to existing computational and biological models} \yoshij{An interesting comparison can be drawn between joint detection-tracking models, including ours, and recently highlighted attention mechanisms. The term attention refers selection mechanisms to extract a useful subset from feature pools~\cite{mnih2014recurrent,bahdanau2014neural}. The attention models currently used can be categorized into two types: soft and hard attention~\cite{xu2015show}. Soft attentions~\cite{bahdanau2014neural,yang2016stacked} compute weighted sums of feature vectors from each location within the image, and the weights of each location adaptively vary. In contrast, hard attentions~\cite{mnih2014recurrent,zaremba2015reinforcement} select only one region at a time; in other words, they assign discrete weights of 0 or 1 to locations, which usually makes the optimization harder. In our framework, the tracking can be regarded as a hard temporal attention mechanism that selects where to look in the following frames. However, a major difference is that ours exploits cross-correlation maps between frames to compute attentions. This makes the usage of hard attention simpler by eliminating the need for stochastic optimization that was necessary in almost all of the existing hard-attention frameworks.}

\yoshij{Digressing from the computational world, motion-induced attention is also seen in  visual nervous systems of animals; thus, our model is biologically plausible. In primates including humans, moving objects cause eye movement to keep the objects' retina images near the fovea; these are called smooth pursuit eye movements~\cite{westheimer1954eye}. The eye movements can be modeled by a negative feedback system that feeds back movements of the objects' retina images and matches the eye movement's velocity to the objects'~\cite{robinson1986model}. In this regard, the RCN's location feedback to search windows can be viewed as an computational analogue of pursuit eye movement.}

% ------------------------------------------------------------------------------------------------
\section{Conclusion}
We introduced the {\it Recurrent Correlation Network}, a novel joint detection and tracking framework \yoshifinal{that exploit motion information of small flying objects}.
In experiments, we tackled two recently developed datasets consisting of images of small flying objects, where the use of multi-frame information is inevitable due to poor per-frame visual information. The results showed that in such situations, multi-frame information exploited by the ConvLSTM and tracking-based motion compensation yields better detection performance. In the future, we will try to extend the framework to multi-class small object detection in videos.

%\begin{acknowledgements}
%\end{acknowledgements}


\bibliographystyle{spmpsci}      % mathematics and physical sciences
\bibliography{bib}   % name your BibTeX data base

\end{document}